\documentclass[journal]{IEEEtran} 
\usepackage[utf8]{inputenc}
\usepackage[printonlyused]{acronym}
\usepackage{xcolor}
\usepackage[font=small]{caption}
\usepackage{subfigure}
\usepackage{array, makecell}
\usepackage{graphicx}
\usepackage{fancyhdr}
\usepackage{balance}
\usepackage{amsmath}
\usepackage{algorithm}
\usepackage{algorithmic}
\usepackage{amsfonts}

\usepackage[maxbibnames=10,style=ieee,backend=bibtex]{biblatex}
\colorlet{blue}{black}
\colorlet{red}{black}

\begin{document}
\title{Neural 3D Object Reconstruction with\\ Small-Scale Unmanned Aerial Vehicles}

\author{Álmos Veres-Vitályos, Filip Lemic, Daniel Johannes Bugelnig, Joan Bernaus Casades\'us,\\ Genís Castillo Gómez-Raya, Sergi Abadal, Bernhard Rinner\IEEEauthorrefmark{1}, Xavier Costa-Pérez
\vspace{-3mm}
\IEEEcompsocitemizethanks{\IEEEcompsocthanksitem\IEEEauthorrefmark{1}Corresponding author.}
\thanks{Á.~Veres-Vitályos, F.~Lemic, J.~Bernaus Casades\'us, G.~Castillo Gómez-Raya, and X.~Costa-Pérez are affiliated with the i2CAT Foundation, Spain, email: \{name.surname\}@i2cat.net.}%
\thanks{D.J.~Bugelnig and B.~Rinner are affiliated with the University of Klagenfurt, Austria, email: \{name.surname\}@aau.at.}%
\thanks{S.~Abadal is affiliated with the Universitat Politècnica de Catalunya, Spain, email: abadal@ac.upc.edu.}%
\thanks{F.~Lemic is also affiliated with the Faculty of Electrical Engineering and Computing, University of Zagreb, Croatia.}%
\thanks{X.~Costa-Pérez is also affiliated with the NEC Labs Europe GmbH, Germany and ICREA, Spain.}%
}

\maketitle

\begin{abstract}
Miniaturized \acp{UAV} can access indoor and hard-to-reach spaces, but severe constraints on payload and autonomy have limited their use in demanding tasks such as high-quality \ac{3D} reconstruction. We introduce a novel system architecture that enables autonomous, high-fidelity 3D scanning of static objects with sub-100~gram UAVs. Our core innovation lies in \textcolor{blue}{a closed-loop active viewpoint selection framework specifically tailored for ultra-constrained micro-platforms, advancing beyond standard static or offline active reconstruction methods. The framework establishes} a dual-reconstruction pipeline that creates a real-time (RT) feedback loop between data capture and flight control. A near-RT process uses Structure-from-Motion (SfM) to generate an instantaneous point-cloud of the object. \textcolor{blue}{A systematic trajectory adaptation algorithm} analyzes the model quality on the fly and dynamically adapts the UAV's trajectory \textcolor{blue}{based on parameterized spatial partitioning} to intelligently capture new images of poorly covered areas, \textcolor{red}{aiming at comprehensive data acquisition}. For the final, high-fidelity output, a non-RT pipeline employs a Neural Radiance Fields (NeRF)-based \ac{GenAI} approach, fusing SfM-derived camera poses with precise \textcolor{blue}{external location data, evaluated across both radio-based \ac{UWB} and visual motion-capture setups, to} \textcolor{red}{mitigate sensor noise and improve accuracy}. We implemented and validated this architecture using Crazyflie 2.1 UAVs. Our experiments, conducted in both single- and \textcolor{red}{dual-UAV configurations utilizing fixed altitude and angular separation,} show that \textcolor{blue}{algorithmic} dynamic trajectory adaptation \textcolor{red}{achieves improvements in key} reconstruction \textcolor{red}{metrics under most experimental conditions} over static flight paths. This work demonstrates a \textcolor{red}{practical} and autonomous solution that unlocks the potential of miniaturized UAVs for fine-grained 3D reconstruction, a capability previously reserved for much larger platforms.
\end{abstract}



\acrodef{KDE}{Kernel Density Estimation}
\acrodef{UAV}{Uncrewed Aerial Vehicle}
\acrodef{SfM}{Structure-from-Motion}
\acrodef{3D}{3-dimensional}
\acrodef{LPD}{Loco Positioning Deck}
\acrodef{LPS}{Loco Positioning System}
\acrodef{UWB}{Ultra Wideband}
\acrodef{TDoA}{Time Difference of Arrival}
\acrodef{VR}{Virtual Reality}
\acrodef{RGB}{Red, Green and Blue}
\acrodef{CLI}{Command Line Interface}
\acrodef{NeRF}{Neural Radiance Fields}
\acrodef{TWR}{Two-Way Ranging}
\acrodef{TDoA}{Time Difference of Arrival}
\acrodef{ULP}{Ultra Low Power}
\acrodef{IoT}{Internet of Things}
\acrodef{IO}{Input-Output}
\acrodef{PCB}{Printed Circuit Board}
\acrodef{GS}{Gaussian Splatting}
\acrodef{CAGR}{Compound Annual Growth Rate}
\acrodef{MLP}{Multilayer Perceptron}
\acrodef{LiDAR}{Light Detection and Ranging}
\acrodef{ToF}{Time of Flight}
\acrodef{AI}{Artificial Intelligence}
\acrodef{2D}{2-dimensional}
\acrodef{3D}{3-dimensional}
\acrodef{instant-ngp}{Instant Neural Graphics Primitives}
\acrodef{GenAI}[N3DR]{Neural 3D Reconstruction}
\acrodef{RT}{real-time}
\acrodef{PSNR}{Peak Signal-to-Noise Ratio}
\acrodef{MSE}{Mean-Squared Error}
\acrodef{SSIM}{Structural Similarity Index Measure}
\acrodef{LPIPS}{Learned Perceptual Image Patch Similarity}
\acrodef{HD}{Hausdorff Distance}
\acrodef{WD}{Wasserstein Distance}
\acrodef{ISM}{Industrial, Scientific, and Medical}
\acrodef{JCAS}{Joint Communications and Sensing}
\acrodef{GPS}{Global Positioning System}
\acrodef{BW}{black-and-white}

\vspace{-1mm}
\section{Introduction}

Recent advances in robotics resulted in an uptake of \acfp{UAV}~\cite{mozaffari2021toward, Rinner_Computer2021}. Additionally, developments in device miniaturization have driven the creation of advanced small \ac{UAV} systems. The development of these small \acp{UAV}, typically weighing less than a kilogram, has been enabled by the miniaturization and cost reduction of electronic components including micro-processors, sensors, batteries, and wireless communication units~\cite{floreano2015science}. Small UAVs are opening novel avenues for applications such as structural monitoring automation~\cite{gordan2021brief}, generation of \ac{VR} content~\cite{talarn2023real}, quality control~\cite{liu2022challenges}, cultural heritage preservation~\cite{campana2017drones}, geology~\cite{giordan2020use}, and more. These \acp{UAV} are primarily envisioned for indoor use, which reduces the need for flight licenses and makes them more appealing for broader market adoption.
Despite this progress, current commercial small \acp{UAV} generally lack the autonomy to navigate without skilled human supervision, which limits the scalability of their missions~\cite{floreano2015science}. \textcolor{blue}{The core challenge is energy: flight is power intensive, and miniaturization exacerbates scaling effects such as lower motor power density, reduced transmission efficiency, and elevated energy costs for maneuvers like  hovering~\cite{mohsan2023unmanned}.} These constraints lead to limitations in flight autonomy, path planning, battery endurance, and payload capability, which hinder the UAV's ability to perform more complex tasks.

Autonomous \acf{3D} reconstruction of static objects is a compelling application for small UAVs~\cite{stathopoulou2019open} and is useful for various domains, including industry~\cite{mourtzis2021uavs}, urban monitoring~\cite{yan2021sampling}, disaster management~\cite{karam2022micro}, warehouse operations~\cite{wawrla2019applications}, subterranean explorations~\cite{de2022rmf}, heritage documentation~\cite{xu2016skeletal}, and \ac{VR}~\cite{maboudi2023review}. While smartphones and handheld devices have enabled progress in \ac{3D} reconstruction, they require manual intervention and cannot be fully automated, limiting scalability. 

In this paper, we demonstrate that small \acp{UAV} can automate this workflow. Specifically, we introduce a system architecture for small UAV-supported autonomous generation of accurate \ac{3D} digital representations of static objects. In our architecture, small \acp{UAV} carry camera sensors and visit a set of locations around the object. They hover to capture images and transmit the data wirelessly to a station running an open-source pipeline for generating a \ac{3D} representation of the object. \textcolor{blue}{Our architecture utilizes near-\ac{RT} reconstruction, rapidly processing image data on the fly to create a closed-loop feedback system that dynamically controls the UAV's flight trajectory and viewpoint selection.} The system also incorporates \acf{NeRF}, a novel \acf{GenAI} approach for accurate 3D reconstruction and volumetric rendering. \textcolor{blue}{The utilization of NeRF-based rendering is advantageous for miniaturized UAV applications, as its loss backpropagation mechanism} \textcolor{red}{can refine} \textcolor{blue}{the noisy camera poses typically acquired by lightweight, low-resolution sensors.}

The key technological innovation of the system lies in its use of sub-100~gram \acp{UAV}, demonstrating how advanced functionality can be achieved \textcolor{blue}{on ultra-light platforms despite severe constraints on payload, power, and onboard computation. This enables truly small-scale, autonomous \ac{3D} reconstruction in unknown, complex, indoor, and hard-to-reach environments where larger systems are impractical.}
\textcolor{blue}{Beyond system integration, the main methodological innovation of the framework stems from its algorithmic approach to active viewpoint selection, establishing a closed-loop between near-\ac{RT} \ac{3D} reconstruction and dynamic \ac{UAV} positioning. Rather than relying on ad-hoc heuristics, the system employs a parameterized spatial partitioning algorithm to systematically assess coverage quality and dynamically generate kinematically feasible waypoints.} This approach enables the \ac{UAV} to adapt its trajectory based on the quality of ongoing spatial reconstruction. 
By leveraging this awareness, the system can intelligently adjust the \ac{UAV}'s path for more accurate reconstructions. 

\begin{figure*}
\centering
  \includegraphics[width=0.78\linewidth]{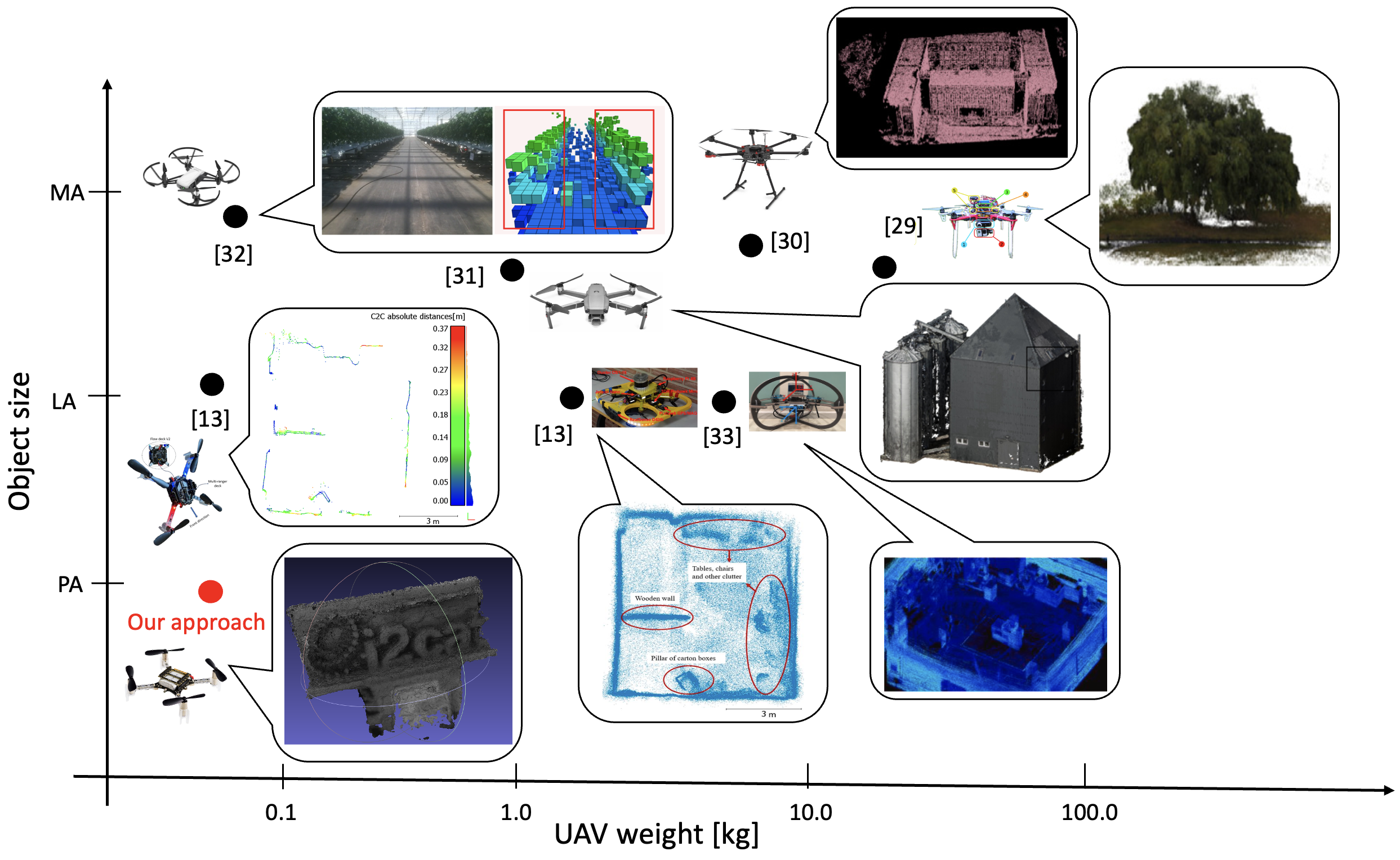}
  \vspace{-1mm}
  \caption{Positioning of the proposed system with regard to the state-of-the-art along two dimensions: UAV weight (kg) vs object size (MA: metropolitan area, LA: large area, PA: personal area).}
  \label{fig:related_efforts}
  \vspace{-3mm}
\end{figure*}

We prototyped the architecture using off-the-shelf Crazyflie micro-\acp{UAV} and an open-source reconstruction stack, including \ac{SfM} for near-\ac{RT} feedback and Nerfacto (via Nerfstudio) for non-\ac{RT} high-fidelity modeling. 
Experimental evaluation demonstrates \textcolor{red}{that dynamic trajectory optimization achieves improvements in several key} 3D reconstruction \textcolor{red}{metrics under most conditions, evaluated} \textcolor{blue}{under both radio-based \acf{UWB} and visual motion-capture localization.}
Compared to existing literature, which typically focuses on larger UAVs and objects (cf., Figure~\ref{fig:related_efforts}), our approach opens new possibilities for fine-grained 3D reconstruction of small objects in constrained environments.

This paper is structured as follows. In Section~\ref{sec:related_works}, we review related work in the field of \ac{UAV}-based \ac{3D} reconstruction and small \ac{UAV} applications. 
Section~\ref{sec:architecture} presents the proposed system architecture, detailing the key components and their interactions. 
In Section~\ref{sec:implementation}, we describe the system implementation, including hardware setup and software pipelines for \ac{3D} object reconstruction. 
Section~\ref{sec:methodology} presents the evaluation methodology, outlining the utilized experimental setup and performance metrics.
In Section~\ref{sec:evaluation_results}, we provide the results of the experiments, including both qualitative and quantitative analyses of the \ac{3D} reconstruction performance. 
Section~\ref{sec:discussion} discusses the findings, and presents directions for future work.
Section~\ref{sec:conclusion} concludes the paper.

\section{Related Works}
\label{sec:related_works}

\subsection{3D Reconstruction of Static Objects}

Photogrammetry extracts \ac{3D} shape information from physical objects~\cite{schenk2005introduction}. Despite its origin in 1867, it remains a cornerstone for 3D object reconstruction across various fields including topographic mapping, archaeology, architecture, infrastructure inspection, and \ac{VR} content generation~\cite{kovanivc2023review}.

Photogrammetry uses two primary data sources: images and \ac{LiDAR} scans~\cite{kovanivc2023review}. LiDAR is a sensing method that employs laser light to measure distances and create precise 3D representations of objects or surfaces. It excels in generating point-clouds by combining depth data with \ac{RGB} imagery. However, LiDAR systems are often too large and heavy which limits their deployment on small UAVs, especially indoors where payload and size are tightly constrained.

Image-based 3D reconstruction are therefore attractive for small-scale UAV applications due to their low cost, energy efficiency, and ability to capture high-resolution textures. One of the most popular image-based methods is \ac{SfM}, which reconstructs 3D structures from overlapping 2D images by identifying matching features across them and estimating camera positions~\cite{kamencay2012improved}. SfM is widely used due to its robustness and ability to handle large datasets, though it can be computationally demanding, particularly during feature matching and optimization stages.

Recent advances in machine learning have further pushed the boundaries of 3D reconstruction, particularly with techniques like \ac{NeRF}~\cite{mildenhall2021nerf, gao2022nerf}, which use neural networks to synthesize novel views of a scene by optimizing its volumetric representation. NeRF is capable of producing high-quality, detailed reconstructions, but its computational requirements can be prohibitive for real-time applications. \ac{GS}~\cite{kerbl3Dgaussians} offers an efficient alternative by approximating the scene with Gaussian functions, enabling faster rendering while maintaining competitive detail levels. Another emerging technique is \ac{instant-ngp}, which leverages a multi-resolution hash table to rapidly reconstruct 3D scenes~\cite{muller2022instant}, providing near-instantaneous results suitable for near-\ac{RT} applications.

We use Nerfacto~\cite{nerfstudio}, a \ac{NeRF}-based \ac{GenAI} implementation, due to its balance between quality and computational feasibility. This choice is supported by our prior work~\cite{castillo2025neural3d}, where we experimentally compared Nerfacto with alternatives like instant-ngp~\cite{muller2022instant} and Splatfacto~\cite{matsuki2024gaussian} on UAV-captured datasets.
The results showed that Nerfacto consistently achieved more accurate and stable reconstructions under the resource and noise constraints of small UAV platforms.

\subsection{UAV-based 3D Object Reconstruction}

\textcolor{blue}{\acp{UAV} are enabling applications across domains such as surveillance, disaster management, and mobile \ac{VR}~\cite{rovira2022review}. The UAV market's projected growth from \$19.3 billion in 2019 to \$45.8 billion by 2025 (15.5\% \ac{CAGR})~\cite{rovira2022review}, underscores their cross-domain impacts.}

\textcolor{blue}{Existing literature on UAV-supported 3D digital reconstruction can be categorized along two dimensions, i.e., target object size and UAV platform weight, as shown in Figure~\ref{fig:related_efforts}. The majority of works focus on outdoor, metropolitan area (MA)-sized targets using mid- to heavy-weight UAV platforms. For instance, multi-kilogram platforms have been deployed for 3D modeling of outdoor scenes and buildings, such as heavy hexacopter setups in~\cite{lienard2016embedded}, 9.5~kg DJI M600 Pro platforms in~\cite{huang2020fast}, and 907~g DJI Mavic Pro 2 systems in~\cite{koch2019automatic}. When targeting indoor MA-scale scenarios like precision agriculture, systems typically rely on monitoring entire farm facilities using dedicated flight spaces~\cite{krul2021visual}.}

\textcolor{blue}{Other efforts target indoor, large area (LA)-sized objects such as residential interiors or industrial facilities. These applications generally employ medium-weight platforms ($\sim$1.45--3.17~kg) equipped with heavy payloads including LiDAR and stereo cameras~\cite{karam2022micro, gao2023uav, de2022rmf}. While micro-UAVs (e.g., Crazyflie) have occasionally been explored for LA indoor mapping~\cite{karam2022micro}, they achieve only rudimentary reconstruction accuracy due to severe payload and processing constraints.}

\textcolor{blue}{Our work addresses the unaligned niche in Figure~\ref{fig:related_efforts} by utilizing ultra-lightweight ($<$100~g) micro-UAVs for high-precision 3D reconstruction of personal area (PA)-sized static objects. Unlike existing efforts that rely on larger platforms or static trajectories, our architecture establishes a closed loop between near-RT SfM coverage assessment and dynamic UAV trajectory adaptation, paired with location-aware NeRF-based rendering (Nerfacto) for high-fidelity volumetric modeling.}

\subsection{Active Viewpoint Selection}
\textcolor{red}{Active viewpoint selection and Next-Best-View (NBV) strategies for 3D reconstruction fall into learning-based active policies, reconstruction-aware control, and offline frame selection. Recently, cooperative multi-agent active mapping frameworks such as COLMAR~\cite{pham2026colmar} utilize multi-agent reinforcement learning (MARL) with overlap-aware objectives to maximize coverage in large environments. Similarly, object-reconstruction-aware whole-body control formulations~\cite{dursun2026object} and dynamic implicit surface mapping~\cite{liu2021active} optimize continuous trajectories for ground-based robots by focusing the sensor field-of-view on low-density object regions. On the algorithmic frame-selection front, combinatorial methods like genetic frame selection~\cite{farchione2026less} demonstrate that active subset filtering improves non-real-time reconstruction quality under tight image budgets compared to uniform sampling.}

\textcolor{red}{While these approaches achieve high fidelity, they rely on depth sensing and a 3D map of the scene, a policy trained in simulation, or a set of already captured images, none of which is available to a sub-100 gram UAV with a low-resolution monocular camera. In contrast, our proposed framework introduces a lightweight, parameterized spatial partitioning formulation. It provides near-real-time dynamic viewpoint selection and multi-UAV spatial segregation (via fixed altitude and angular bounds) tailored to the energy, payload, and processing limits of micro-aerial platforms.}


\section{System Architecture}
\label{sec:architecture}

\begin{figure*}
\centering
  \includegraphics[width=0.92\linewidth]{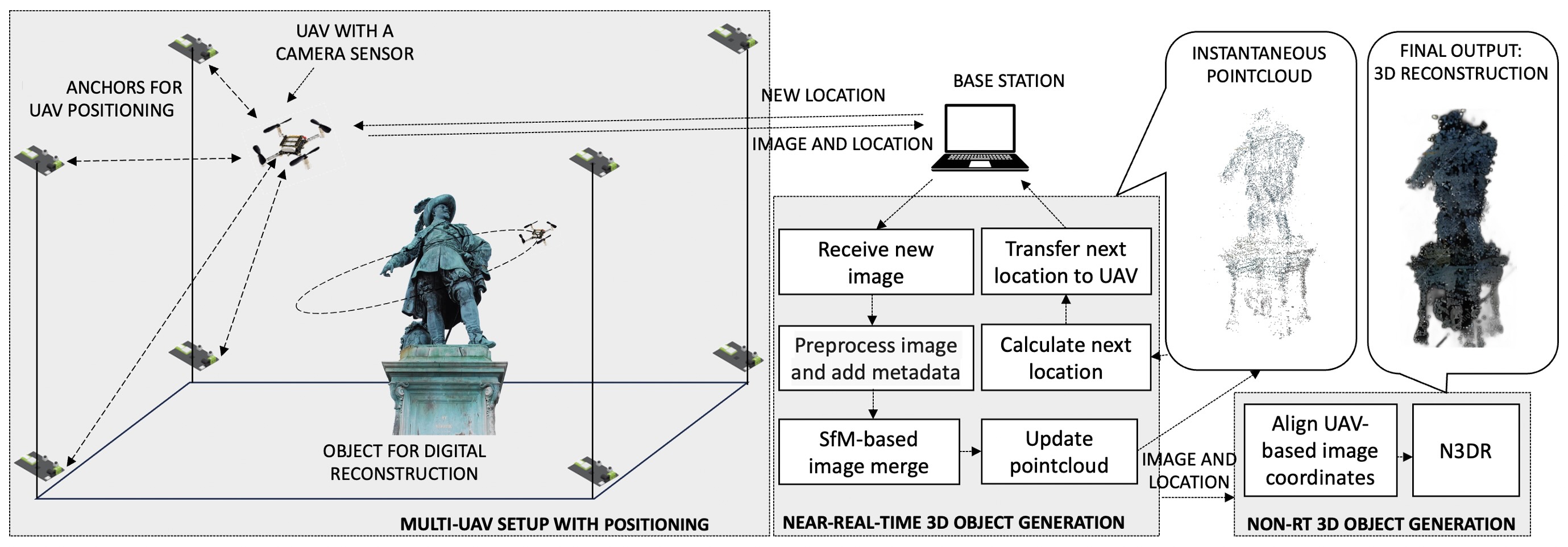}
  \vspace{-1mm}
  \caption{Our system architecture is composed by UAVs supported by a positioning system and a base station executing near-RT and non-RT 3D object generation. The UAVs circle around the object and transfer images from predefined locations to the base station. The images are merged to an instantaneous point-cloud in near-RT and together with location data to the 3D reconstruction in non-RT. The current coverage level of the object is estimated with the instantaneous point-cloud, and a new location in \textcolor{red}{the region with the lowest coverage score} is sent to the UAVs (dynamic trajectory adaptation).}
  \vspace{-3mm}
  \label{fig:setup}
\end{figure*}

Figure~\ref{fig:setup} depicts the proposed system architecture for generating \ac{3D} object representations. The architecture consists of three components: (i) the multi-\ac{UAV} setup with the positioning support and the base station containing (ii) the near-\ac{RT} and (iii) non-\ac{RT} generation pipelines.

The setup consists of UAVs equipped with onboard camera sensors and a supporting positioning system for their localization. The UAVs circle the object along fixed waypoints, capture images from various angles, and transfer images with location data to the base station. The base station can adapt the trajectory by sending new locations to the UAVs.  
The near-RT generation component processes the images received from the UAVs in real time. The images are preprocessed and merged into a rough 3D model of the object (instantaneous point-cloud), which is used to estimate the coverage level of the object, enabling the UAVs to adjust their waypoints to improve the reconstruction quality.
The non-RT reconstruction component creates the high-fidelity 3D model using non-RT techniques. This process takes advantage of the data gathered during the near-RT processing and employs advanced methods like \acp{NeRF} to produce detailed volumetric rendering.

Our system is initialized with the waypoints of the initial circles around the object, and the UAVs need to face the object before take-off. To initiate 3D reconstruction, the base station triggers the take-off and movement to the initial location. 
The UAVs then follow to the predefined or dynamically computed locations depending on whether static or dynamic trajectories have been selected. If no further locations are provided, the reconstruction mission is completed, and the UAVs descend for landing. Our system can handle a variable number of UAVs, and in this paper we present results with one and two UAVs.

To bootstrap the reconstruction, at least two images with a small but nonzero baseline are required. Accordingly, each \ac{UAV} first captures two front-facing images of the object from slightly offset positions to introduce parallax. A near-\ac{RT} method, such as \ac{SfM}, then generates an initial point-cloud.

Each image is accompanied by metadata containing the camera parameters (i.e., focal length, sensor size, image size, camera serial number), which are later used for 3D reconstruction. 
The images are preprocessed, for example, by removing the background, applying a sharpening filter, or adjusting the brightness. This image processing depends on the particular application to increase the effectiveness of feature detection in images and the accuracy of the estimated camera positions.

Each metadata-annotated and preprocessed image is sequentially processed in the near-\ac{RT} pipeline with the core steps of obtaining the intrinsic camera parameters (from image metadata), feature extraction and matching, and applying \ac{SfM}. 
The camera positions and poses for each image are estimated based on the camera intrinsic parameters, which is easy to utilize for the reconstruction, yet it could feature low estimation accuracy especially for low-quality images taken by small \acp{UAV}.

\subsection{Dynamic UAV Trajectory Adaptation}

The dynamic trajectory adaptation guides the \ac{UAV} towards viewpoints that improve the reconstruction quality of \textcolor{red}{the least-covered} object areas, while avoiding redundant image capture. 
\textcolor{red}{To achieve this, the UAV's circular flight path around the object is divided into $R$ regions (also referred to as slices), which are fixed angular segments of the flight circle, defined by the azimuth of the UAV's camera orientation relative to the center of the flight path. Each region corresponds to a specific viewing direction towards the object and serves as the basic unit for coverage evaluation.}

\begin{figure*}[!t]
  \centering
  \subfigure[SfM-originating locations]{
  \includegraphics[width=0.234\textwidth]{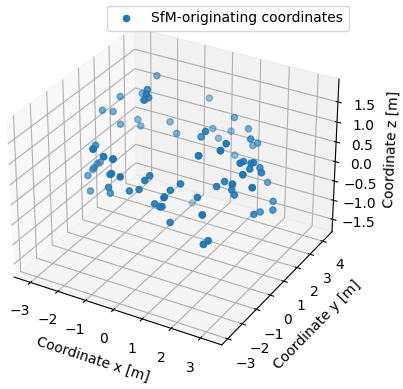}}
  \subfigure[UAV-originating locations]{
  \includegraphics[width=0.234\textwidth]{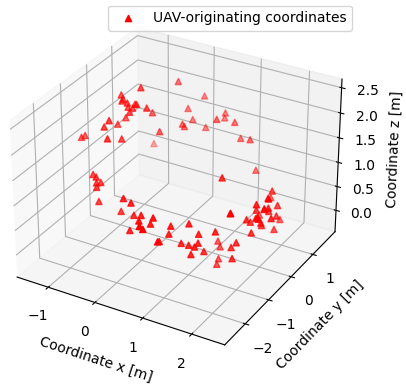}}
  \subfigure[Alignment of UAV locations]{
  \includegraphics[width=0.234\textwidth]{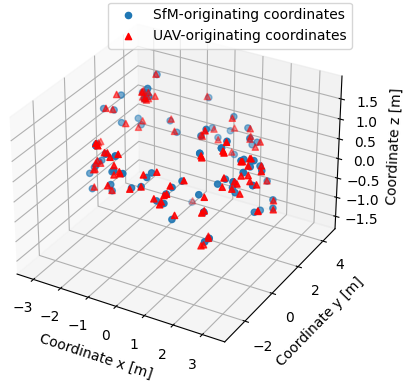}}
  \subfigure[Final locations]{
  \includegraphics[width=0.234\textwidth]{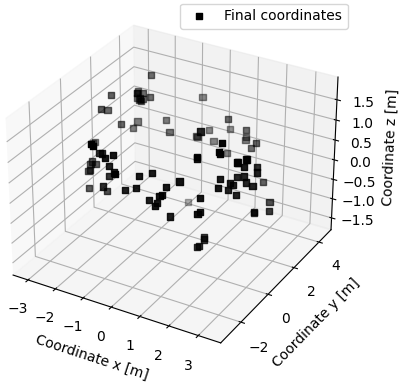}}
  \vspace{-1mm}
  \caption{Alignment of \ac{SfM}- and \ac{UAV}-originating image locations illustrated for a representative experiment using static, circular trajectories around the object. The depicted camera positions obtained through \ac{SfM} (a) are generated by default in the reconstruction process. The positions obtained from the UAV positioning system (b) are jointly utilized with the \ac{SfM}-derived positions in the location-aware non-RT reconstruction. This integration (c) involves appropriate scaling, rotation, and alignment to ensure consistency between the two coordinate frames before serving as input to the N3DR model (d).}
  \vspace{-4mm}
  \label{fig:loco_colmap}
\end{figure*}


\textcolor{red}{Within each region, the distance of every point to its fifth nearest neighbor is sorted, and Knee-Point detection on the lowest $\tau_{cut}=0.9$ fraction of the sorted curve yields the DBSCAN radius $\epsilon$, which avoids a hardcoded clustering radius. Points with a larger neighbor distance are discarded and DBSCAN is applied. The largest cluster and the cluster nearest to the camera positions of the region are retained, and the midpoint $\hat{C}_k$ of their means is taken as the object center of the region. A spherical filter is then applied to the raw points:}
\begin{equation}
\begin{split}
\mathcal{P}_{\text{sphere},k} &= \left\{ P_i \in \mathcal{P}_{\text{raw},k} \;\middle|\; \|P_i - \hat{C}_k\|_2 \le R_k \right\}, \\ 
R_k &= \|\hat{C}_k - \bar{X}_k\|_2,
\end{split}
\end{equation}
\textcolor{red}{where $\bar{X}_k$ is the mean \ac{SfM}-estimated camera position of region $k$ ($R_k$ is expressed in \ac{SfM} units and recomputed in every iteration). The binary indicator $I_{\text{sphere}}$ equals 1 if the number of points inside the sphere exceeds half of the size of the largest cluster and lies within $80$--$120\%$ of the size of the nearest cluster, and 0 otherwise (the largest cluster is then used); it is a count-consistency check rather than a geometric confirmation of the object. Up to two further DBSCAN passes (skipped if $\epsilon < 0.2$ or fewer than 50 points remain)~\cite{bugelnig2026adaptive} yield the filtered point set $\mathcal{P}_{\text{filtered},k}$.}

\textcolor{blue}{The reconstruction quality is then quantified using a coverage score $S_k$, computed} \textcolor{red}{as a weighted linear combination of the point counts (divided by $N_{\text{ref}} = 1000$, but not capped), the final clustering radius $\epsilon_{\text{norm}}$ (capped at $1.0$; a smaller value indicates a denser point-cloud), and $I_{\text{sphere}}$, with $w_{\text{c}} = 0.2$ and $w_{\text{s}} = 0.3$. Since the counts are not capped, a very dense region can obtain a high score irrespective of the structural terms. The score is computed as:}
\begin{equation}
\textcolor{red}{S_k = w_{\text{c}} \, \frac{|\mathcal{P}_{\text{raw},k}| + |\mathcal{P}_{\text{filtered},k}|}{N_{\text{ref}}} + w_{\text{s}} \left[ (1 - \epsilon_{\text{norm}}) + I_{\text{sphere}} \right].}
\end{equation}
\textcolor{blue}{The UAV’s next waypoint $(x_{\text{uav}}, y_{\text{uav}}, z_{\text{uav}})^{\top}$ is then determined} \textcolor{red}{by ranking all regions by their coverage score after every reconstruction iteration, without an absolute coverage threshold, and} \textcolor{blue}{set as the geometric center of the region with the lowest coverage score $\arg\min_k S_k$.} 
\textcolor{blue}{This logic is formalized in Algorithm~\ref{alg:trajectory}.}

\textcolor{blue}{To satisfy UAV kinematic and yaw angle constraint, trajectory generation ensures the object remains continuously centered within the camera by locking the UAV's yaw angle $\psi$ at any waypoint toward the target centroid. Furthermore, during the near-RT processing, the UAV does not pause its flight, but continues to slow-orbit or hover along the circular boundary, ensuring continuous operation until the next waypoint is received.}
\textcolor{blue}{
In multi-UAV configurations, collision avoidance is ensured through strict spatial segregation. Specifically, concurrent UAVs are launched at opposing positions and assigned to a fixed relative altitude offset ($\Delta z = 10\text{ cm}$).}

\subsection{Location-aware Non-RT 3D Reconstruction with N3DR}

The estimation and tracking of \ac{UAV} locations are needed for UAV control. We fuse location information from the near-\ac{RT} photogrammetry pipeline and the positioning system as shown in Figure~\ref{fig:loco_colmap}. 
To improve reconstruction accuracy, we combine the UAV's available coordinates with those generated by the near-RT photogrammetry pipeline. 
This combination reduces errors in estimating the UAV's exact location when each image is taken. 
\textcolor{red}{The alignment between the SfM-derived and UAV-originating trajectories is a single global transformation, which does not compensate for the camera offset from the localization origin. No explicit calibration of the camera pose relative to the localization origin was performed in our implementation, and that the approach assumes an identical camera mounting across the fleet. Residual yaw-dependent pose errors are thus expected.}

\begin{algorithm}[!t]
\vspace{1mm}
\captionsetup{font=small}
\caption{Active viewpoint selection and trajectory adaptation}
\label{alg:trajectory}
\footnotesize
\begin{algorithmic}[1]
\STATE \textbf{Initialize:} Set $R$ angular regions \textcolor{red}{and image budget $B$}
\WHILE{\textcolor{red}{image budget $B$ is not exhausted}}
    \STATE \textcolor{red}{Assign captured images to regions by camera azimuth and height}
    \FOR{$k = 1$ \TO $R$}
        \STATE \textcolor{red}{Execute near-RT SfM on images of region $k$ to obtain $\mathcal{P}_{\text{raw},k}$}
        \STATE \textcolor{red}{Compute Knee-Point $\epsilon$; discard sparse points; run DBSCAN}
        \STATE \textcolor{red}{Select largest and nearest cluster; estimate $\hat{C}_k$ and $R_k$}
        \STATE \textcolor{red}{Apply spherical filter (Eq.~(1)); compute $I_{\text{sphere}}$}
        \STATE \textcolor{red}{Refine with up to two DBSCAN passes to obtain $\mathcal{P}_{\text{filtered},k}$, $\epsilon_{\text{norm}}$}
        \STATE \textcolor{blue}{Compute coverage score} \textcolor{red}{$S_k$ using Eq. (2)}
    \ENDFOR
    \STATE $k_{\text{next}} \leftarrow \arg\min_k(S_k)$ \textcolor{red}{over regions with remaining viewpoints}
    \STATE Dispatch next waypoint $(x_{\text{uav}}, y_{\text{uav}}, z_{\text{uav}})^{\top}$ to center of region $k_{\text{next}}$
    \STATE \textcolor{red}{Maintain UAV yaw angle towards the object center}
\ENDWHILE
\end{algorithmic}
\end{algorithm}

\section{System Implementation}
\label{sec:implementation}

\subsection{Multi-UAV Setup and Localization Systems}

\begin{figure}
\centering
\includegraphics[width=0.98\columnwidth]{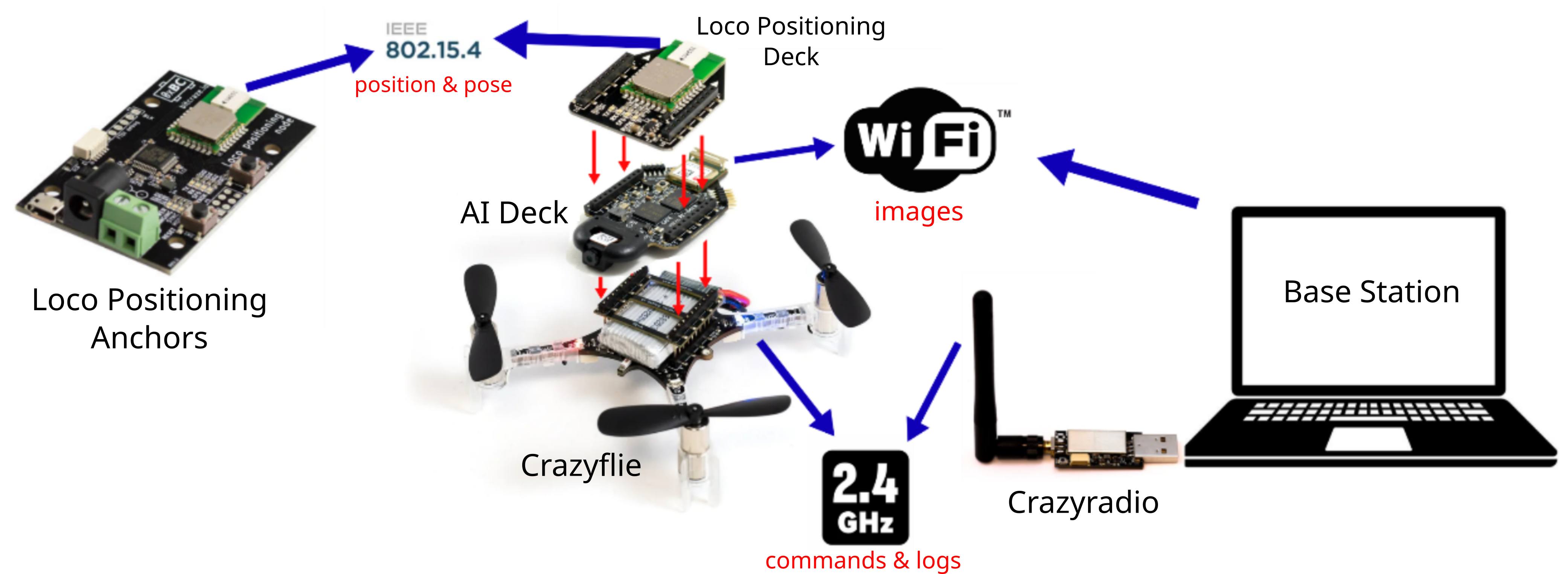}
  \vspace{-1mm}
  \caption{Hardware components and communication interfaces of the implemented system architecture.}
  \label{fig:hardware_setup}
\vspace{-3mm}
\end{figure}

\begin{figure}[!t]
\centering
\subfigure[Loco UWB-supported setup]{
\includegraphics[width=0.23\textwidth]{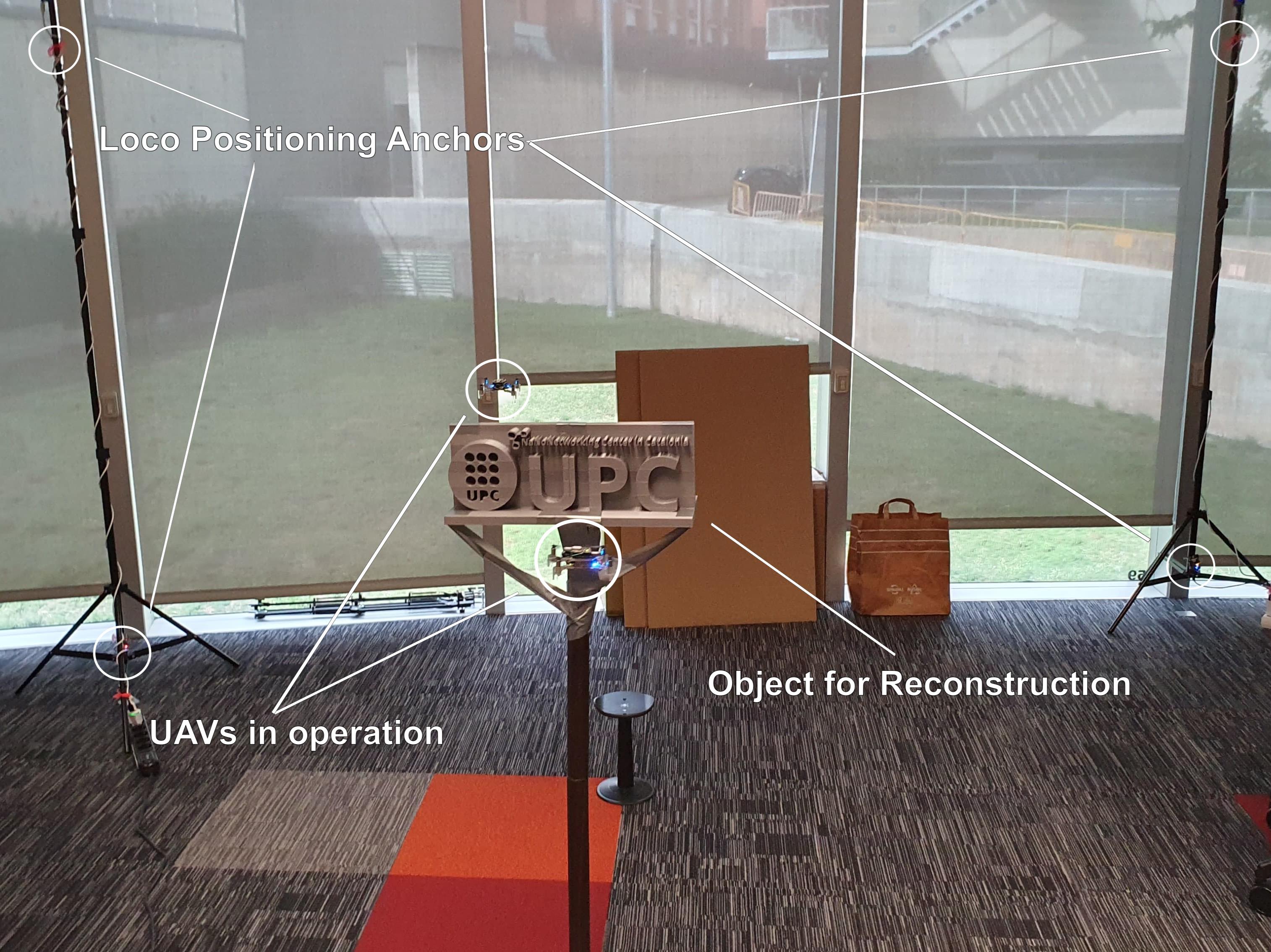}\label{fig:setup_loco}}
\subfigure[OptiTrack-supported setup]{
\includegraphics[width=0.23\textwidth]{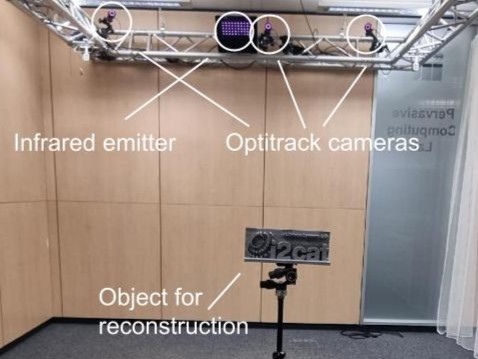}\label{fig:setup_optitrack}}
\vspace{-1mm}
\caption{Deployment in an indoor laboratory setting showing the object for reconstruction within each localization volume. (a) Loco UWB radio localization utilizing anchor stands for sub-decimeter tracking. (b) OptiTrack optical motion-capture localization utilizing an overhead infrared cameras for sub-millimeter ground-truth tracking.}
\label{fig:implemented_system_new}
\vspace{-3mm}
\end{figure}

Figure~\ref{fig:hardware_setup} depicts the selected hardware components for the implementation of the architecture. We use Crazyflie~2.1 \acp{UAV} equipped with a proprietary Crazyradio dongle operating at the 2.4~GHz ISM band for control communication with the base station. Each Crazyflie is equipped with an AI-Deck module hosting a GAP8 processor for efficient image processing, an ultra-low-power camera (320$\times$320 pixels), and an ESP32 Wi-Fi module. The captured images are transferred to the base station using the integrated GAP8 CPU and the Wi-Fi module.
\textcolor{blue}{To evaluate the framework under different localization conditions, the system supports two positioning infrastructures:}
\begin{itemize}
    \item \textcolor{blue}{\textbf{Loco UWB (radio-based localization):} As shown in Figure~\ref{fig:setup_loco}, this setup uses the \ac{LPS} operating in \ac{TDoA} mode where each \ac{UAV} estimates its position relative to fixed anchors. The measured distances are Kalman filtered by the onboard STM32 microcontroller to provide sub-decimeter positioning accuracy suitable for flexible, infrastructure-light deployments.}
    \item \textcolor{blue}{\textbf{OptiTrack (visual motion-capture):} As shown in Figure~\ref{fig:setup_optitrack}, this system operates in a dedicated capture volume equipped with 8 infrared cameras, providing sub-millimeter ground-truth positioning and orientation tracking to benchmark upper-bound pose fusion performance.}
\end{itemize}

\textcolor{blue}{The reliance on external localization systems rather than onboard visual odometry is driven by the strict payload and computational limits of sub-100g micro-\acp{UAV}. Running complex monocular visual odometry algorithms requires memory and processing capabilities that exceed the bounds of the onboard GAP8 microcontroller, which is heavily utilized for image capture and Wi-Fi streaming.}
Figure~\ref{fig:implemented_system_new} shows the system deployed in our laboratories.

\begin{figure*}[!t]
\centering
\subfigure[BW image of the small object]{
\includegraphics[width=0.27\textwidth]{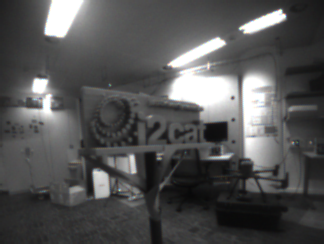}}
\subfigure[Preprocessed BW image]{
\includegraphics[width=0.27\textwidth]{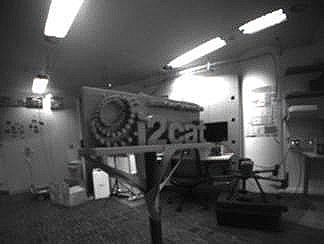}}
\subfigure[RGB image of the large object]{
\includegraphics[width=0.205\textwidth]
{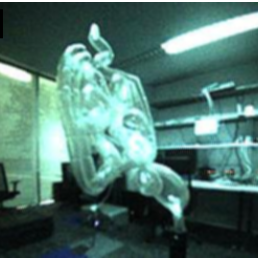}}
\subfigure[Preprocessed RGB image]{
\includegraphics[width=0.205\textwidth]{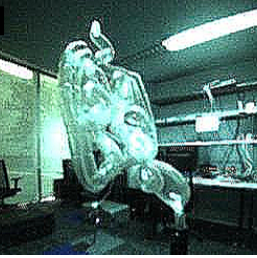}}
\vspace{-2mm}
\caption{Preprocessing examples for BW and RGB images, applied in both near- and non-RT reconstructions.}
\vspace{-4mm}
\label{fig:image_editing}
\end{figure*} 

\subsection{Near-RT 3D Object Generation Pipeline}

The trajectory planning for each Crazyflie involves circling the object at a predetermined distance, either following a static trajectory based on predefined coordinates or dynamically adjusting the flight trajectory based on near-\ac{RT} feedback.
The \ac{LPS} \textcolor{blue}{and OptiTrack systems provide} accurate positional data, which is crucial for maintaining the desired flight path and ensuring consistent object coverage from multiple angles.

\begin{figure*}
\centering
\begin{minipage}{.625\textwidth}
  \includegraphics[width=\linewidth]{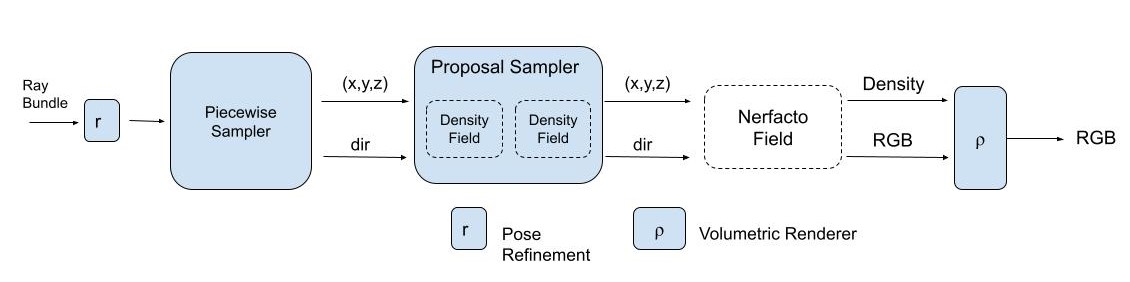}
  \vspace{-2mm}
  \captionof{figure}{Nerfacto pipeline.}
  \label{fig:nerfactoP}
  \vspace{-2mm}
\end{minipage}%
\hfil
\begin{minipage}{.323\textwidth}
  \centering
  \includegraphics[width=\linewidth]{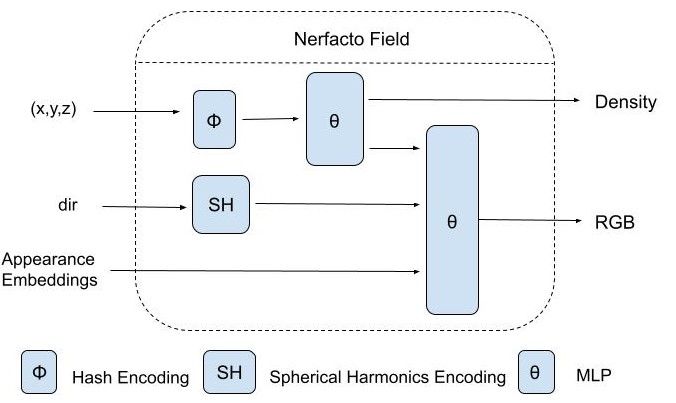}
  \vspace{-4mm}
  \captionof{figure}{Nerfacto Field architecture.}
  \label{fig:nerfactoF}
\end{minipage}%
  \vspace{-5mm}
\end{figure*}

The near-\ac{RT} processing pipeline starts with the reception of images. 
Regardless of the object and camera type, the images are preprocessed by a sharpening filter and brightness correction to enhance visual features and contrast (cf., Figure~\ref{fig:image_editing}).
Specifically, the \textit{Unsharp Masking} and high-pass filters were used for sharpening to improve edge definition and fine details. 
For contrast enhancement, \textit{Histogram Equalization} and \textit{Gamma Correction} were employed to adjust brightness and improve visibility. These methods were selected empirically based on their effectiveness in the given context, although additional optimizations could further refine image quality.

Images are then merged on the \ac{UAV}'s yaw and timestamp, ensuring coherent reconstruction of the object's surface and structure. \textcolor{red}{DBSCAN-based spatial clustering is employed per region to refine the point-cloud.}
\textcolor{blue}{The processing is distributed across four dedicated cores on the base station to optimize computational efficiency and enable near-RT operation during flight. To determine the configuration that best balances reconstruction coverage resolution with system latency, the spatial partitioning is systematically parameterized.}

\subsection{Non-RT 3D Object Generation Pipeline}

The open source Meshroom software from AliceVision~\cite{alicevision2021} was used for implementing the non-RT 3D generation pipeline. The SfM-originating camera positions are used in addition to \ac{LPS} \textcolor{blue}{or OptiTrack} coordinates. This is done by combining the output of the \ac{SfM} and merging it with the rotated and translated Crazyflie-originating coordinates, as indicated in Figure~\ref{fig:loco_colmap}. The implementation of this location-aware approach is carried out in the form of a Python script, which identifies the images used in the estimated reconstruction, retrieves the corresponding location data, and applies a coordinate transformation to merge it with the SfM-originating coordinates. \textcolor{red}{Note that no calibration of the camera pose relative to the UAV's center of mass (the origin point for localization) was performed. Since the camera offset is fixed in the UAV body frame, its direction in the world frame rotates with the UAV's yaw, and a single global similarity transformation between the SfM and positioning frames cannot represent it. 
The resulting yaw-dependent position error may be partially mitigated by the SfM-derived poses and the pose refinement of Nerfacto.}

Figure~\ref{fig:nerfactoP} depicts the implementation of the \ac{GenAI}-based non-RT reconstruction functionality using Nerfacto~\cite{nerfstudio}. This pipeline follows a standard approach for \ac{GenAI}, utilizing the default optimizations provided by the Nerfacto model. 
In a comparative analysis conducted in a previous study~\cite{castillo2025neural3d}, we evaluated several N3DR models and found that Nerfacto consistently delivered high accuracy in 3D reconstruction. 

Errors in the projected camera poses are expected, especially for mobile cameras. Misaligned stances can cause a decrease in sharpness and clarity as well as hazy artifacts. The system may backpropagate loss gradients to the input pose calculations thanks to the NeRF pipeline, particularly with the \textbf{Pose Refinement}. With this knowledge, improving and polishing the position estimates is possible.

The \textbf{Piecewise Sampler} creates the first set of scene samples. Up to a distance of one meter from the camera, this sampler evenly distributes half of the samples. Every sample that remains is distributed so that the step size increases. The frustums are scaled replicas of themselves when the step size is selected. Far-off items can be handled while maintaining a dense set of data for close-by objects by raising the step sizes.
The sample locations are combined by the \textbf{Proposal Sampler} to the areas of the scene that are most important to the final render (usually the first surface intersection). This significantly raises the caliber of 3D reconstruction. The density function for the scene is needed by the proposal network sampler. There are several ways to construct the density function, but it was empirically found that the smallest fused \ac{MLP} with a hash encoding works well and quickly. The proposal network sampler can be chained together with different density functions to further consolidate the sampling. Two density functions are preferable to one~\cite{nerfstudio}. 

To direct sampling, the \textbf{Density Field} has to depict a rough density representation of the scene. A quick technique to query the scene is to combine a tiny fused \ac{MLP} from tiny-cuda-nn with a hash encoding. We increase its efficiency by reducing the size of the encoding dictionary and the number of feature levels. 
\textbf{Nerfacto Field} is used for rendering different types of visualizations, including the NeRF-based render, point-cloud or mesh visualization, and the rendered video.
Figure~\ref{fig:nerfactoF} depicts the Nerfacto Field architecture with two types of encodings: the Hash encoding and a spherical harmonics encoding with the ray direction as input. Finally, the \ac{MLP} generates the density and the images.

The density field is used to figure out if there is an object on that part of the scene, while the \ac{RGB} is utilized to generate the color of that part of the scene. Once training is completed, the overall density field and the \ac{RGB} field are used to compose the NeRF visualization, that then renders new views of the scene.
After generating and rendering the different views, a point-cloud is obtained by using the samples of the different rays that come from the viewpoint and pixels of each image, while a mesh is acquired as a Poisson surface reconstruction from the normals of the models~\cite{poissond2006Kazhdan}.

\vspace{-1mm}
\section{Evaluation Methodology}
\label{sec:methodology}

This section provides an overview of the evaluations performed to assess the 3D reconstruction performance of the proposed system. The evaluations are based on images captured by small UAVs with location data from \textcolor{blue}{both Loco UWB-based and OptiTrack optical motion-capture positioning}, tested under single- and multi-UAV setups with static and dynamic trajectories, and using both \ac{BW} (i.e., grayscale) and \ac{RGB} (i.e., color) images. A set of heterogeneous metrics is used to assess the reconstruction quality.

\begin{figure*}[!t]
\centering
\subfigure[Small reference object]{
\includegraphics[width=0.21\textwidth]{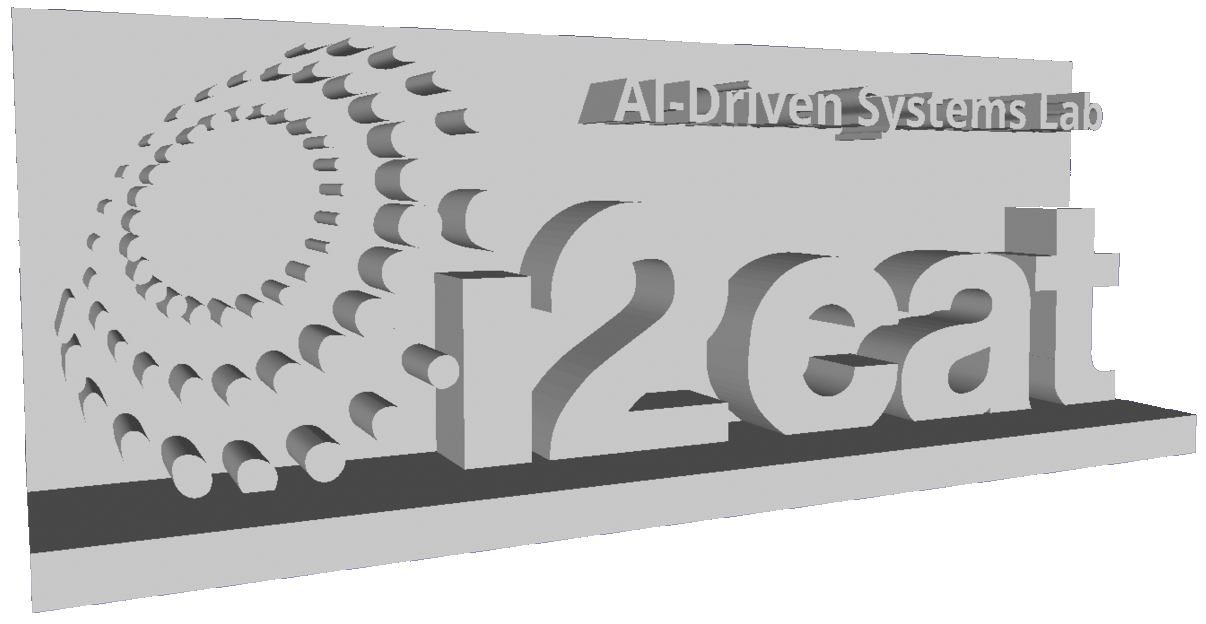}}
\subfigure[Small object's backside]{
\includegraphics[width=0.21\textwidth]{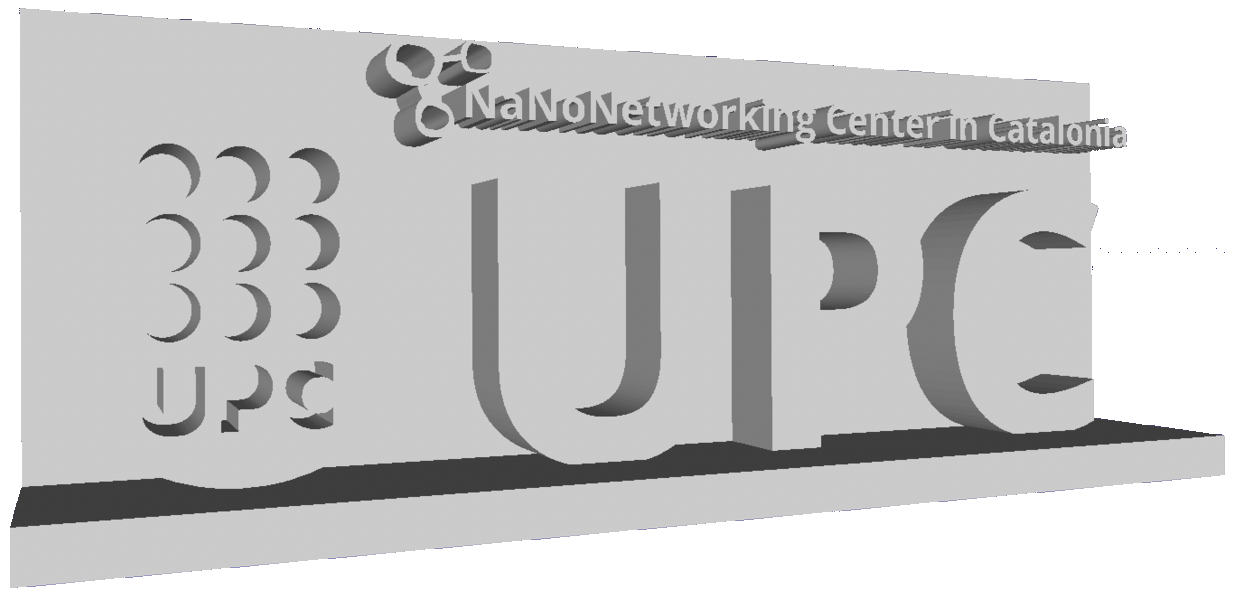}}
\subfigure[\hspace{-0.85mm}Large~object]{
\includegraphics[width=0.089\textwidth]{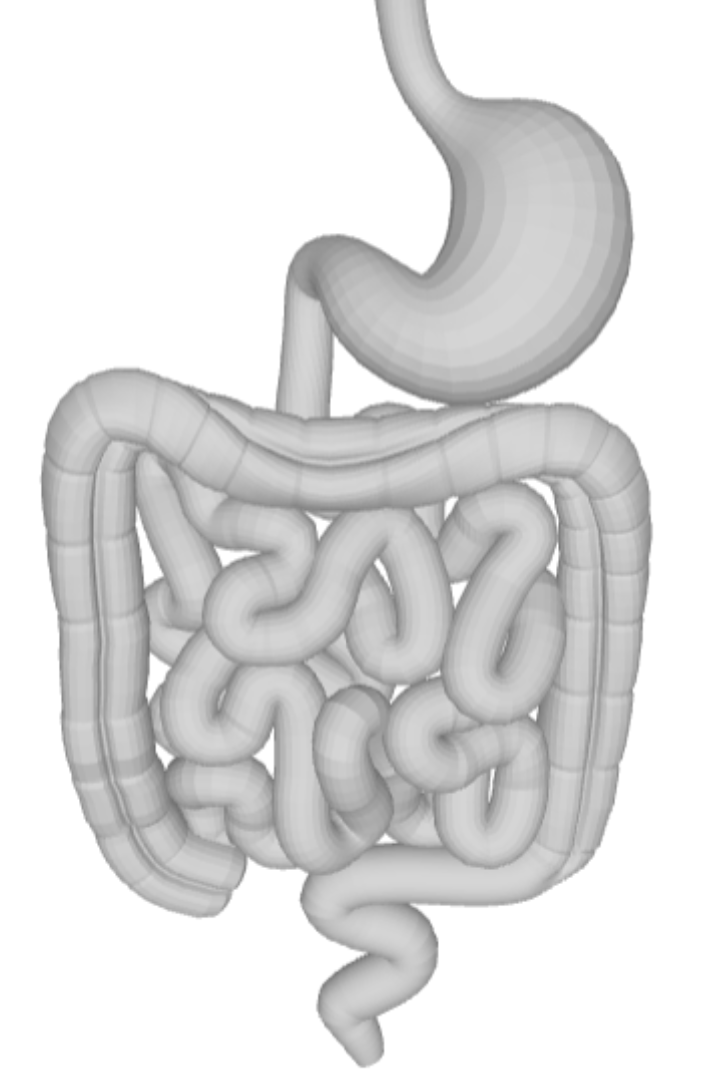}}
\subfigure[Nerfacto render]{
\includegraphics[width=0.21\textwidth]{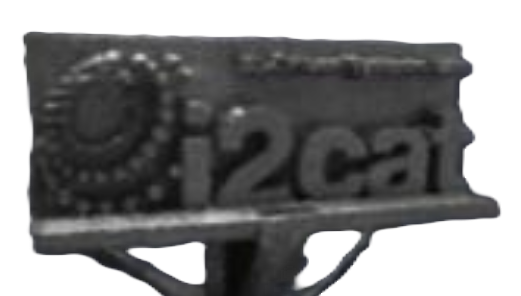}}
\subfigure[Mesh visualization]{
\includegraphics[width=0.21\textwidth]{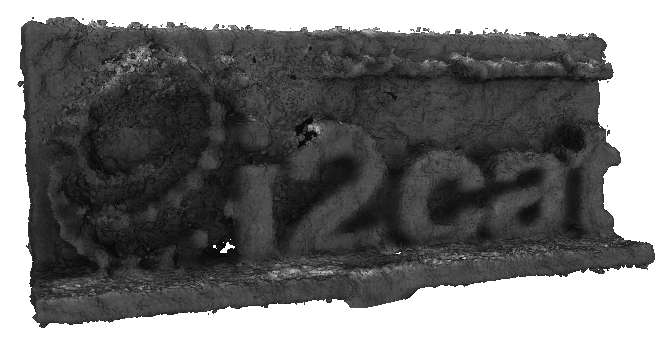}}
\vspace{-1mm}
\caption{Object for reconstruction and resulting visualizations}
\vspace{-4mm}
\label{fig:object}
\end{figure*} 

\vspace{-1mm}
\subsection{Evaluation Scenarios}
\label{subsec:evaluation_scenarios}

The small reference object is depicted in Figure~\ref{fig:object}a-b.
The resulting 3D-printed object has a size of 54.7$\times$20.3$\times$20.9~cm\textsuperscript{3} with letters and engravings 4~cm depth.
Figure~\ref{fig:image_editing}a shows an example \ac{BW} image of this object captured by the \ac{UAV}. 

To evaluate the system’s scalability and robustness under more complex visual and structural conditions, we introduce a 3D-printed anatomical model of the human gastrointestinal system as a second reference object. This object, depicted in Figure~\ref{fig:object}c, is fabricated using transparent polymer, which introduces significant challenges for visual reconstruction due to light refraction and low-contrast internal features.
Moreover, its anatomical shape and vertical span of approximately 90~cm, represent a substantial departure from the compact, high-contrast first object. 
This test case is used to assess the limits of both the near- and non-RT reconstruction pipelines when facing large, visually ambiguous geometries.
Figure~\ref{fig:image_editing}c depicts an example \ac{RGB} image of this object. 

For the static experiments, the UAVs followed a static trajectory around the objects maintaining an approximate distance of \textcolor{red}{50~cm from each object}. The objects were positioned at a height of approximately 1 m from the floor. Each UAV was able to complete 2-3 circles around the object, capturing approximately 250 images in the process. In the dual-UAV experiments, the UAVs were initially positioned opposing each other and followed the same static flight trajectories, with a small difference of 10 cm in altitude during concurrent circles.
Since related works deployed more powerful \acp{UAV} and/or targeted larger objects (cf., Section~\ref{sec:related_works}), we used  the static experiment as baseline and evaluated the impact of both location-awareness and dynamic path adaptation as new system functionalities.

\textcolor{blue}{To evaluate the framework across different positioning operational limits, experiments were executed using Loco UWB radio positioning and OptiTrack motion-capture positioning.
As both positioning systems operate in distinct physical volumes and coordinate spaces, results are evaluated within their respective experimental series rather than directly compared.}

\textcolor{blue}{To validate system functionalities and assess parameter sensitivity, a four-dimensional ablation study was conducted under the high-precision OptiTrack setup. First, the fleet size was varied between single and dual \ac{UAV} configurations. Second, spatial partitioning was assessed across $R\in\{2, 4, 6, 8\}$ regions. Third, the pipeline phase was isolated (near-\ac{RT} \ac{SfM} vs. non-\ac{RT} \ac{NeRF}). Fourth, the pose source was evaluated by comparing pure \ac{SfM}-derived camera poses against OptiTrack-fused location-aware poses.}
Experiments were conducted using \ac{BW} and RGB camera input modes to evaluate the influence of color information on reconstruction fidelity. Moreover, each configuration was tested using both a near-RT pipeline and a non-RT offline NeRF-based pipeline to assess trade-offs between reconstruction quality and processing time.

\textcolor{red}{For the image-budget and OptiTrack parameter analyses,} \textcolor{blue}{to evaluate the trajectory adaptation strategies under repeatable conditions and isolate the impact of viewpoint selection, data acquisition was conducted exhaustively across the viewing volume.
The dynamic trajectory strategies were subsequently evaluated by actively subsampling viewpoints from this pre-captured pool.} \textcolor{red}{During these replays, the selection algorithm analyzes the images it has selected up to the current iteration. The image closest to a dispatched waypoint is taken from the pool, regions whose pool of images is exhausted are skipped, and the replay ends when the image budget is exhausted.}

\textcolor{red}{In contrast, the experiments targeting radio-based localization performance, scalability, and latency were obtained from autonomous online closed-loop flights, in which the UAVs generated near-RT point-clouds and updated their waypoints mid-flight.}

The result of each experiment is the \ac{3D} reconstructed object in the form of a Nerfacto render (cf., Figure~\ref{fig:object}d), as well as mesh (cf., Figure~\ref{fig:object}e) and point-cloud visualizations. We base the evaluation on point-clouds, given their broad adoption and the availability of libraries for computing the selected performance metrics.
To ensure meaningful evaluation, alignment between reconstructed point-clouds and a reference was critical.
Point-cloud orientation and scaling consistency were achieved by computing scaling factors from bounding box ratios and employing fast global registration and principal component analysis for orientation correction. 
Alignment was refined through global registration techniques based on feature matching and iterative closest point algorithm.
Post-alignment, virtual cameras were generated \textcolor{red}{randomly} around the object and directed towards its center, to render images from both the reference and reconstructed point-clouds. 
Performance metrics were subsequently calculated directly from the point-clouds or from comparisons with the rendered images.

\begin{figure*}
\centering
\begin{minipage}{.491\textwidth}
    \centering
    \subfigure[Baseline]{
    \includegraphics[width=0.478\linewidth]{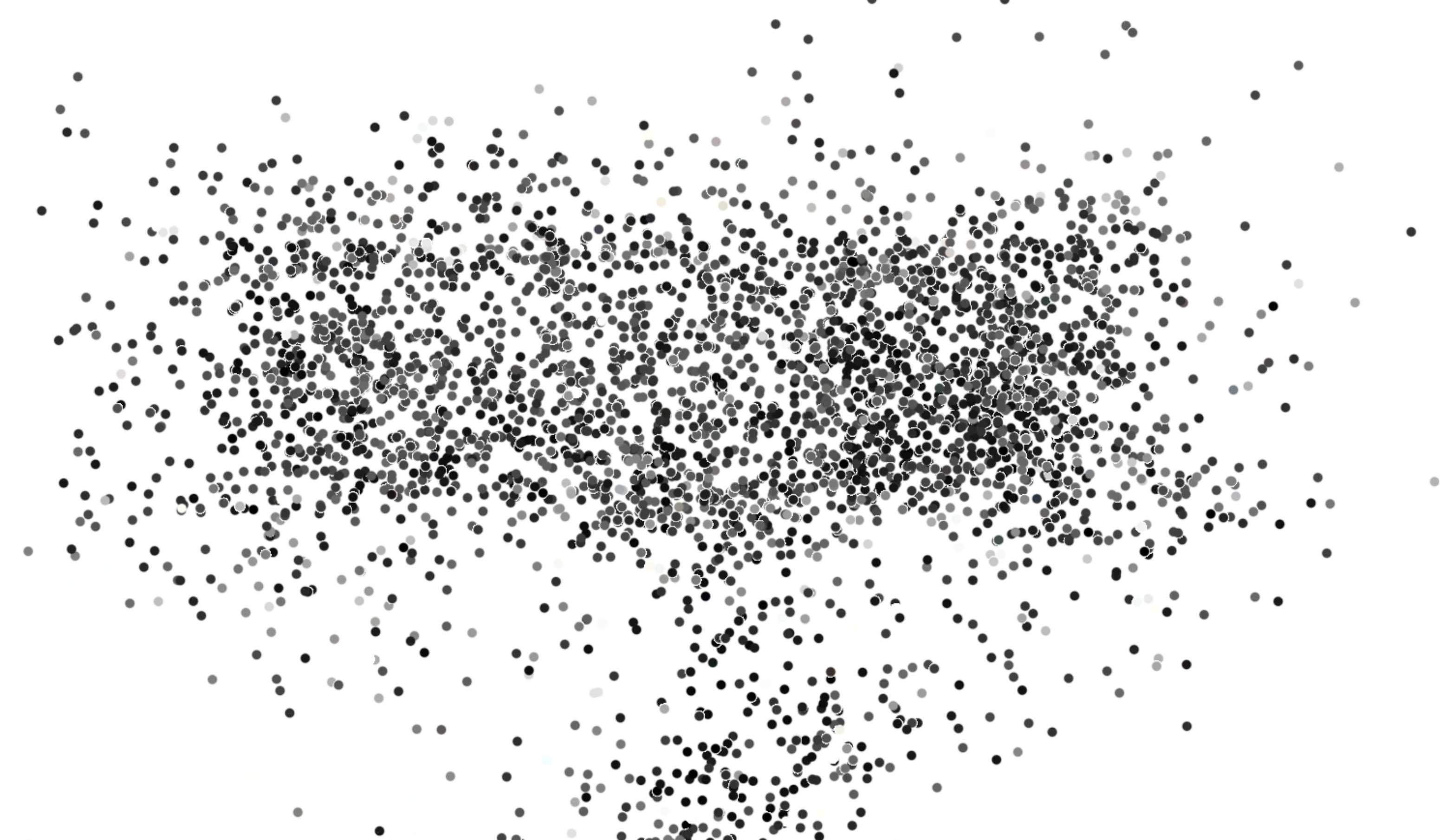}}
    \subfigure[Dynamic path]{
    \includegraphics[width=0.478\linewidth]{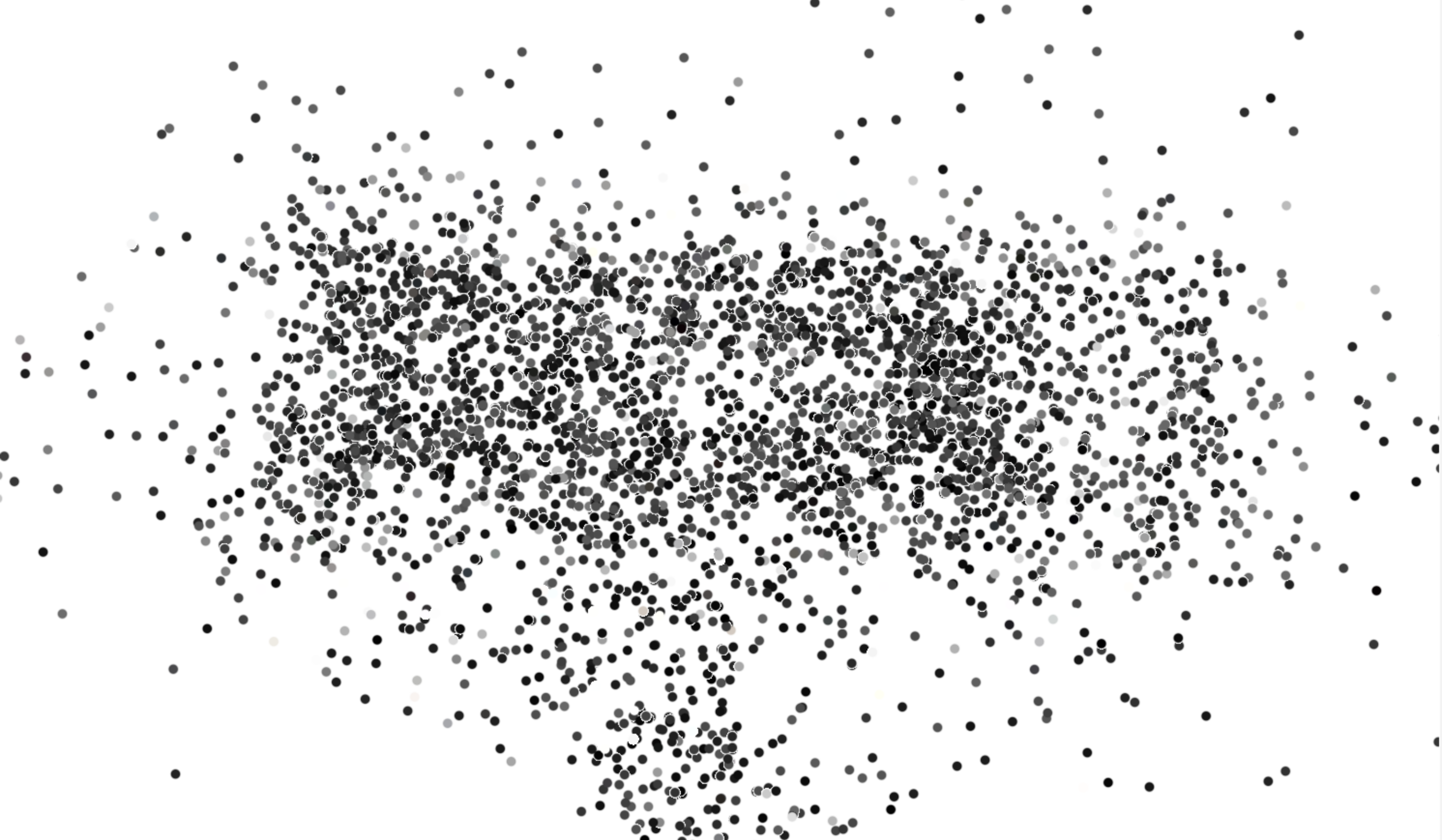}}
  \captionof{figure}{Near-\ac{RT} reconstruction for single-UAV setup.}
  \label{fig:visualization_results}
\end{minipage}%
\hfil
\begin{minipage}{.49\textwidth}
  \centering
    \centering
    \subfigure[Baseline]{
    \includegraphics[width=0.478\textwidth]{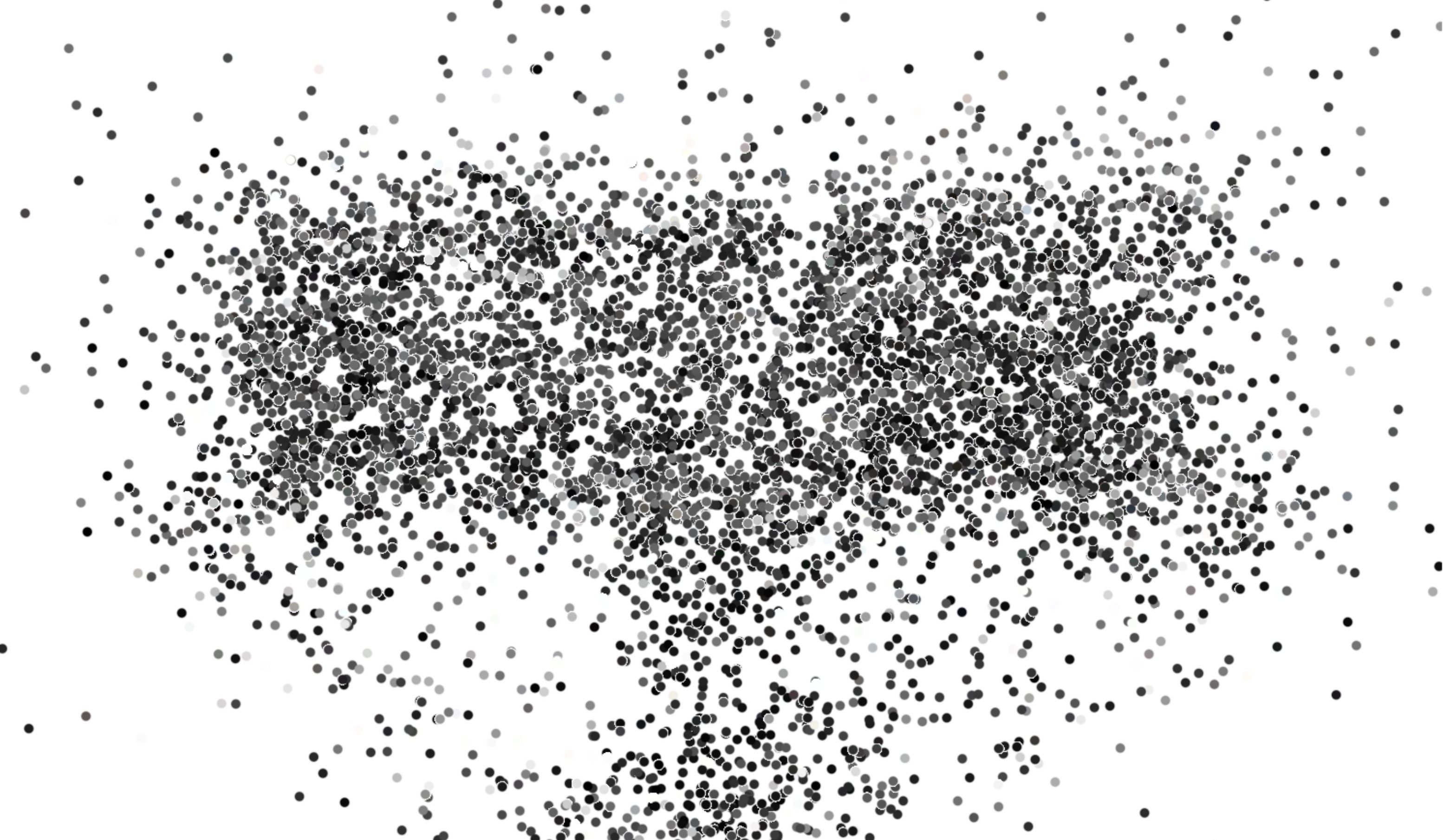}}
    \subfigure[Dynamic path]{
    \includegraphics[width=0.478\textwidth]{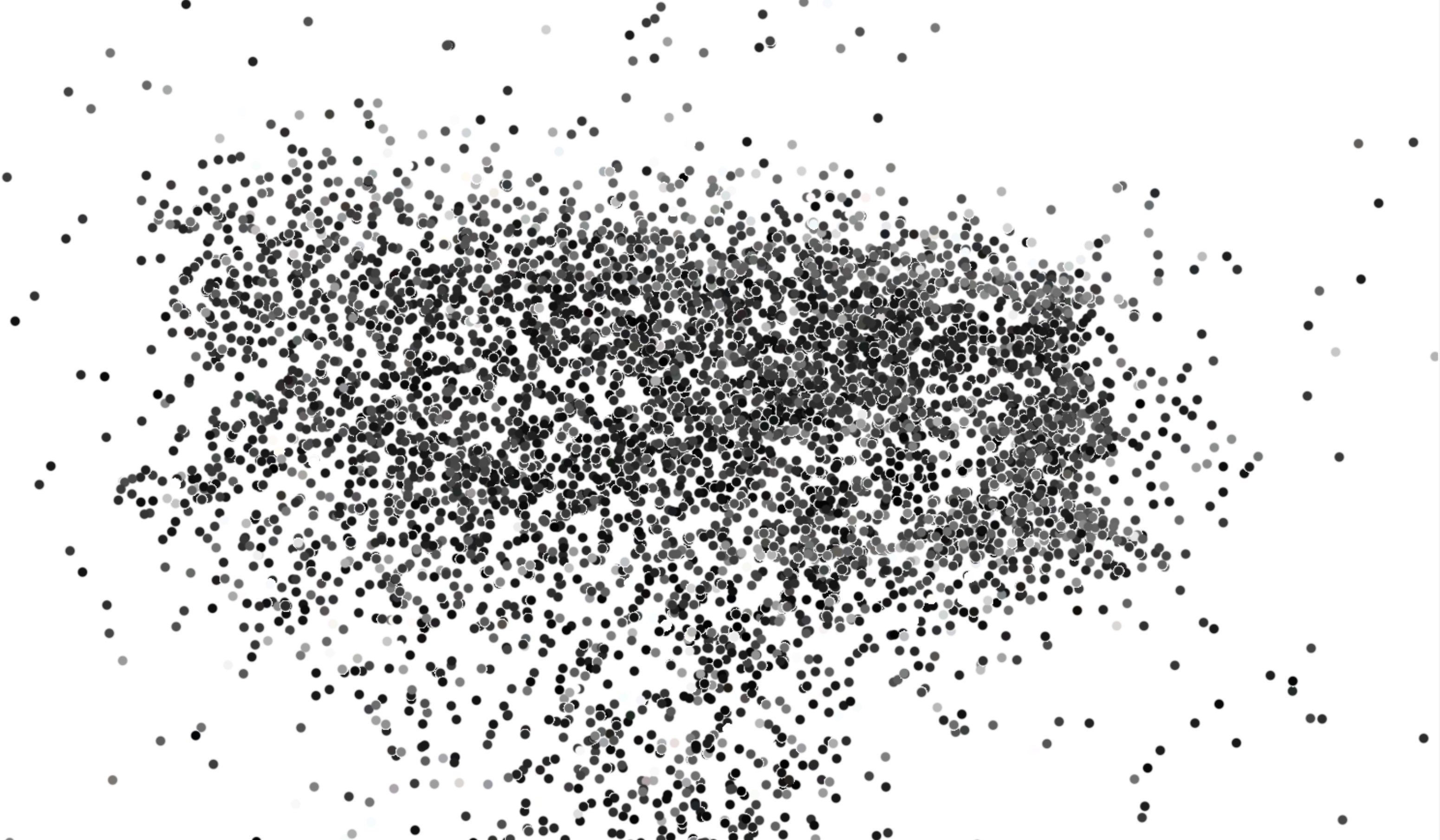}}
  \captionof{figure}{Near-\ac{RT} reconstruction for dual-UAV setup.}
  \label{fig:visualization_results_2}
\end{minipage}
\vspace{-3mm}
\end{figure*}

\begin{figure*}[!t]
\centering
\subfigure[Baseline]{
\includegraphics[width=0.236\textwidth]{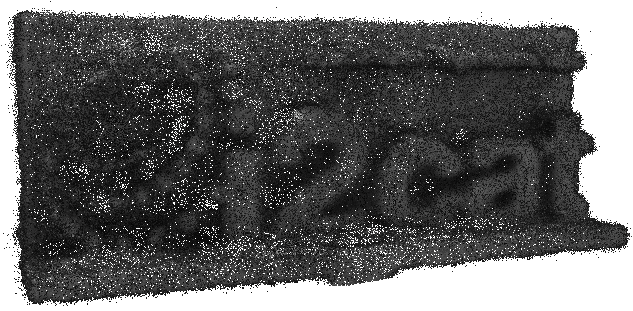}}
\subfigure[Location-aware]{
\includegraphics[width=0.236\textwidth]{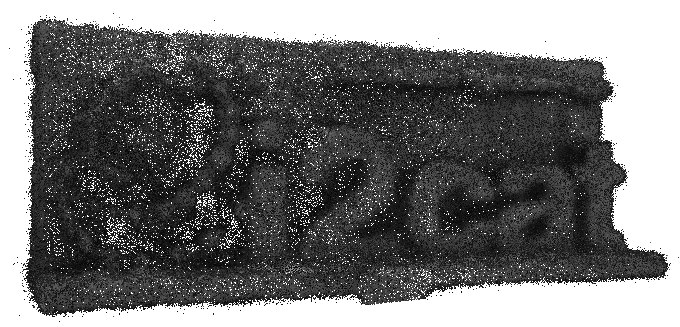}}
\subfigure[Dynamic path]{
\includegraphics[width=0.236\textwidth]{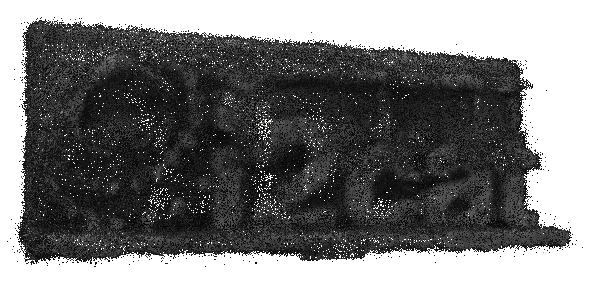}}
\subfigure[Integrated]{
\includegraphics[width=0.236\textwidth]{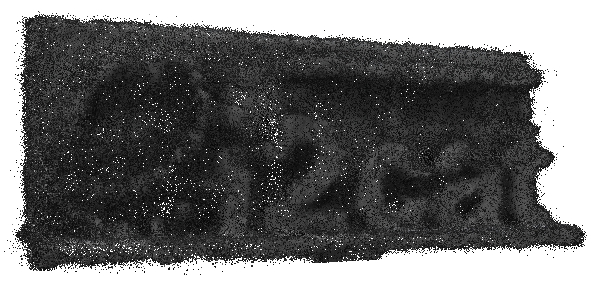}}
\vspace{-1mm}
\caption{Non-\ac{RT} reconstruction for single-UAV setup.}
\vspace{-4mm}
\label{fig:visualization_results_nrt}
\end{figure*} 

\begin{figure*}[!t]
\centering
\subfigure[Baseline]{
\includegraphics[width=0.236\textwidth]{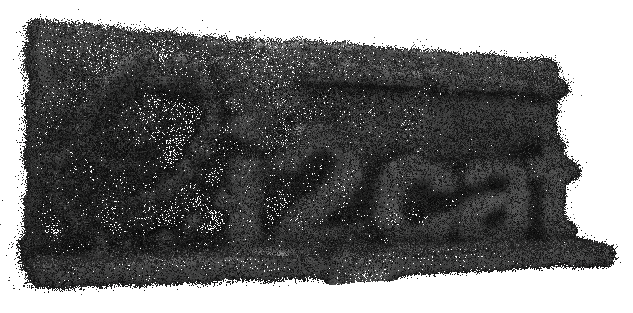}}
\subfigure[Location-aware]{
\includegraphics[width=0.236\textwidth]{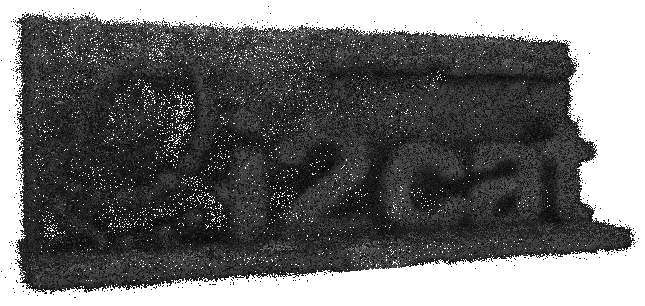}}
\subfigure[Dynamic path]{
\includegraphics[width=0.236\textwidth]{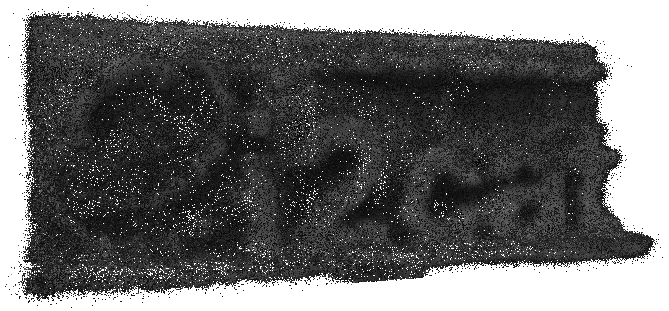}}
\subfigure[Integrated]{
\includegraphics[width=0.236\textwidth]{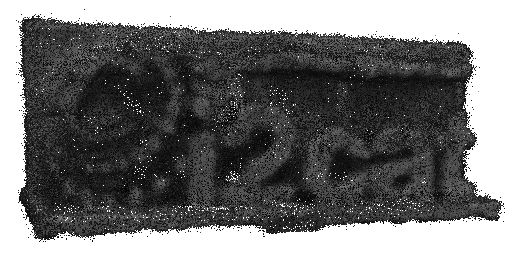}}
\vspace{-1mm}
\caption{Non-\ac{RT} reconstruction for dual-UAV setup.}
\vspace{-3mm}
\label{fig:visualization_results_nrt_2}
\end{figure*} 

\vspace{-1mm}
\subsection{Performance Metrics}

We use the following performance metrics to assess the quality of the point-clouds.
\textbf{\ac{PSNR}} measures the ratio between the maximum possible signal power and the noise affecting its quality~\cite{hore2010image}. 
\textbf{\ac{SSIM}} evaluates the structural similarity between the images and the 3D render. It yields a value between -1 and 1, with 1 indicating perfect similarity and 0 indicating no similarity. The metric combines luminance, contrast, and structure comparisons across different image windows~\cite{wang2003multiscale}.

\textbf{\ac{LPIPS}} measures perceptual similarity between images and the 3D render. A value of 0 is optimal, while values approaching 1 indicate poorer similarity. This metric matches human perception closely by comparing the activations of image patches within a neural network~\cite{zhang2018unreasonable}.
\textbf{\ac{HD}} is the maximum distance between points in the reference and reconstructed point-clouds, indicating overall reconstruction deviation. 
\textbf{\ac{WD}} calculates the minimum work required to transform the reconstructed point-cloud into the reference. It is effective for comparing point-clouds by considering their underlying probability distributions.

In the near-RT reconstruction, the \textbf{reconstruction latency} measures the time taken to merge a new image set with the existing pointcloud. In the non-RT scenario, this latency refers to the total time required for NeRF training and rendering.
Finally, we report the \textbf{number of images captured} by the UAVs and subsequently used in each reconstruction experiment, as an indicator of trajectory coverage and system scalability.
\textcolor{red}{Higher values of PSNR and SSIM, and lower values of LPIPS, HD, WD, and latency indicate better performance; this is marked by $\uparrow$ and $\downarrow$ in the tables.}

\section{Evaluation Results}
\label{sec:evaluation_results}

\textcolor{red}{Our experimental campaign uses two evaluation modes: online closed-loop flights and offline replays. The experiments on radio-based localization (Section~\ref{subsec:eval_loco}), scalability (Section~\ref{subsec:eval_scalability}), and latency (Section~\ref{subsec:eval_latency}) were obtained from fully autonomous online missions in which the UAV generated near-RT point-clouds and updated its waypoints mid-flight. The image budget and parameter sensitivity analyses (Sections~\ref{subsec:eval_budget} and~\ref{subsec:eval_optitrack}) replay exhaustively pre-collected image datasets.
Since the two modes use different image data, results are compared only within each mode, and absolute metric values are not compared across modes.}

\subsection{Performance under Radio-Based Localization}
\label{subsec:eval_loco}

This set of experiments focuses exclusively on the reconstruction of the small reference object using \ac{BW} images \textcolor{blue}{within the Loco UWB radio localization volume.} \textcolor{blue}{The goal is to evaluate the real-world operational impact of location-awareness and dynamic path planning under realistic sub-decimeter tracking noise.}

As shown in Figures~\ref{fig:visualization_results} and~\ref{fig:visualization_results_2}, the near-\ac{RT} reconstruction is hardly recognizable to the human eye. However, it delivers a rough shape and contours of the object, regardless of the utilized approach. This level of fidelity is sufficient for \ac{UAV} control tasks such as dynamic trajectory adaptation for reconstruction quality optimization. This is exemplified in the dynamic path approach, where the \ac{UAV} positioning is performed in a way that optimizes the quality of the reconstructed object. This dynamic path adaptation results in \textcolor{red}{an overall favorable improvement trend in the} quality of near-\ac{RT} reconstruction, as detailed in Tables~\ref{tab:results_reconstruction_single_uwb} and~\ref{tab:results_reconstruction_dual_uwb} for single- and dual-UAV setups.

\begin{table*}[!t]
  \caption{Single-UAV 3D reconstruction results (Loco UWB) \textcolor{red}{comparing location-aware reconstruction and dynamic trajectory adaptation against the baseline.}}
  \label{tab:results_reconstruction_single_uwb}
    \centering
    \footnotesize
    \begin{tabular}{l c c c c c c c c}
    \hline
        \textbf{Approach} & \makecell{\textbf{PSNR} \textcolor{red}{$\uparrow$} \\\textbf{[dB]}} &  \textbf{SSIM} \textcolor{red}{$\uparrow$} &  \textbf{LPIPS} \textcolor{red}{$\downarrow$} &  \textbf{HD} \textcolor{red}{$\downarrow$} &  \textbf{WD} \textcolor{red}{$\downarrow$} & \makecell{\textbf{Latency} \textcolor{red}{$\downarrow$} \\\textbf{[sec]}} & \makecell{\textbf{\# images} \\\textbf{taken}} & \makecell{\textbf{\# images} \\\textbf{used}}  \\ \hline
        \multicolumn{9}{c}{\textbf{Near-\acl{RT} 3D reconstruction}} \\ 
        Baseline &3.295 &$0.888\pm0.025$ &$0.2490\pm0.039$ &0.6324 &0.0249 & $23.044\pm9.04$&233 &232 \\ 
        Dynamic path &4.095 &$0.886\pm0.027$ & $0.2764\pm0.053$ &0.5733 &0.0275 & $23.648\pm9.99$ &271 &266 \\ \hline 
        \multicolumn{9}{c}{\textbf{Non-\acl{RT} 3D reconstruction}} \\ 
        Baseline & 7.720 & $0.961\pm0.006$& $0.0429\pm0.008$ & 0.0686& 0.0208& 224 &233 &232 \\  
        Location-aware & 7.817& $0.962\pm0.007$ &$0.0408\pm0.007$ &0.0668 &0.0219& 237&233 &233 \\ 
        Dynamic path & 7.639&$0.961\pm0.008$ &$0.0419\pm0.008$ & 0.0742& 0.0215& 240&271 &271 \\ 
        Integrated & 7.723&$0.962\pm0.005$ &$0.0411\pm0.007$ &0.0726 &0.0207& 257&271 & 271 \\ 
        \hline
    \end{tabular}
    \vspace{-1mm}
\end{table*}

\begin{table*}[!t]
  \caption{Dual-UAV 3D reconstruction results (Loco UWB).}
  \label{tab:results_reconstruction_dual_uwb}
    \centering
    \footnotesize
    \begin{tabular}{l c c c c c c c c}
    \hline
        \textbf{Approach} & \makecell{\textbf{PSNR} \textcolor{red}{$\uparrow$} \\\textbf{[dB]}} &  \textbf{SSIM} \textcolor{red}{$\uparrow$} &  \textbf{LPIPS} \textcolor{red}{$\downarrow$} &  \textbf{HD} \textcolor{red}{$\downarrow$} &  \textbf{WD} \textcolor{red}{$\downarrow$} & \makecell{\textbf{Latency} \textcolor{red}{$\downarrow$} \\\textbf{[sec]}} & \makecell{\textbf{\# images} \\\textbf{taken}} & \makecell{\textbf{\# images} \\\textbf{used}}  \\ \hline
        \multicolumn{9}{c}{\textbf{Near-\acl{RT} 3D reconstruction}} \\ 
        Baseline & 4.651& $0.901\pm0.018$& $0.2255\pm0.028$&0.3991 &0.0240 &$34.756\pm21.14$ & 466 &461 \\ 
        Dynamic path & 5.624&$0.927\pm0.009$ &$0.1572\pm0.018$ &0.3447 &0.0239 &$31.225\pm13.16$ & 507&501 \\ \hline
        \multicolumn{9}{c}{\textbf{Non-\acl{RT} 3D reconstruction}} \\ 
        Baseline  & 7.651&$0.963\pm0.006$ &$0.0397\pm0.007$ &0.0686 &0.0219 & 222 &466 &465 \\  
        Location-aware &7.695 &$0.965\pm0.007$ &$0.0375\pm0.006$ &0.0704 &0.0210 & 217 & 466& 466 \\ 
        Dynamic path  &7.704 &$0.963\pm0.006$ &$0.0404\pm0.008$ &0.0685 &0.0227 & 232 &507 &507 \\ 
        Integrated &7.561 &$0.964\pm0.005$ &$0.0390\pm0.008$ &0.0725 &0.0226 & 226 & 507 & 507 \\ 
        \hline
    \end{tabular}
    \vspace{-3mm}
\end{table*}

\begin{figure*}[!t]
\centering
\includegraphics[width=0.94\textwidth]{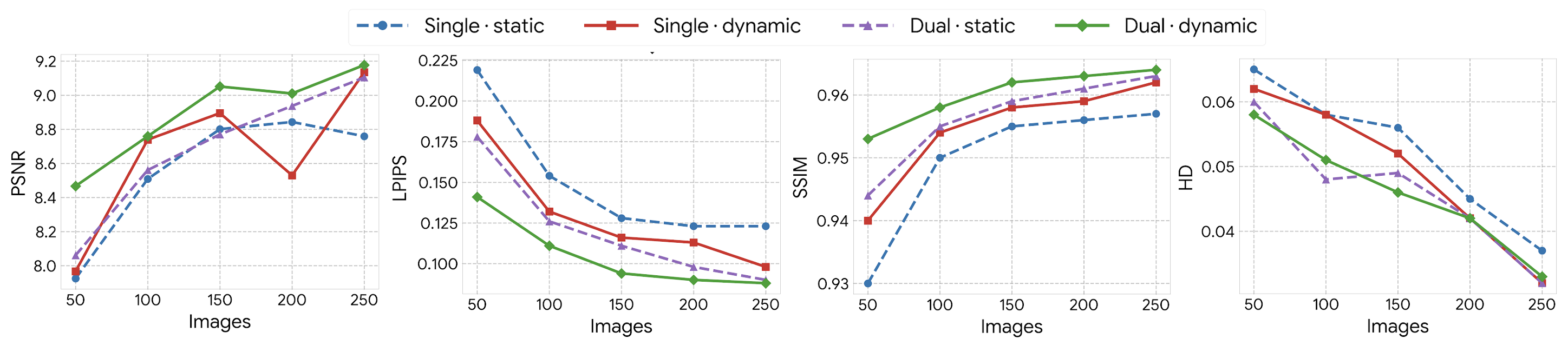}
\vspace{-1mm}
\caption{\textcolor{blue}{Parameter sensitivity analysis evaluating the impact of constrained image budgets on non-RT reconstruction metrics (PSNR, SSIM, LPIPS, and HD) across single- and dual-UAV configurations.}}
\label{fig:plot_per_budget}
\vspace{-2mm}
\end{figure*}

As shown by the quantitative metrics in Tables~\ref{tab:results_reconstruction_single_uwb} and~\ref{tab:results_reconstruction_dual_uwb}, the utilization of location-awareness and dynamic trajectory adaptation improves overall reconstruction quality \textcolor{red}{under most configurations}. Beyond subtle visual refinements (cf., Figures~\ref{fig:visualization_results_nrt} and~\ref{fig:visualization_results_nrt_2}), \textcolor{red}{these approaches lower the near-RT geometric error (HD) in both setups and the non-RT perceptual error (LPIPS) in five of six configurations (the non-RT HD does not improve systematically),} while also increasing the total number of images successfully utilized by the reconstruction pipeline. \textcolor{red}{However, the gains differ across metrics, given that in the single-UAV near-RT case, PSNR and HD improve, whereas SSIM, LPIPS, and WD do not (Table~\ref{tab:results_reconstruction_single_uwb}), possibly because the spatial partitioning constraints occasionally prioritize macro-geometric coverage over the localized dense sampling required for high-fidelity texture rendering.}

Specifically, the location-aware approach increases the reliability of the point-cloud generation as it provides an additional source of camera locations. In other words, obtaining coordinates utilizing \ac{SfM} is possible only if the aligned images have sufficient features for camera localization, while in the location-aware approach the \ac{UAV} can be utilized as a substitute. This eventually results in a larger number of images used by the N3DR approach for reconstruction. \textcolor{blue}{Tables~\ref{tab:results_reconstruction_single_uwb} and~\ref{tab:results_reconstruction_dual_uwb}} show this effect, where the location-aware approaches effectively leverage the spatial coordinates captured by the \acp{UAV}.

In addition, the dynamic path approach does not \textcolor{red}{substantially} degrade the performance of non-\ac{RT} reconstruction compared to the baseline, although its primary optimization objective is the near-\ac{RT} reconstruction accuracy. \textcolor{blue}{However, some inconsistencies were observed in the dual-UAV integrated approach, which combines dynamic path adaptation with location-awareness, under UWB localization (e.g., a slight PSNR drop in Table~\ref{tab:results_reconstruction_dual_uwb}\textcolor{red}{, from 7.651~dB (baseline) to 7.561~dB (integrated)}). The effect of pose fusion under high-precision localization is examined separately in Section VI-C.}

\begin{table*}[!t]
  \caption{\textcolor{blue}{Comprehensive ablation study results under OptiTrack motion-capture localization, demonstrating the impact of fleet size, spatial partitioning, pipeline phase, and pose source (SfM vs. OptiTrack) on 3D reconstruction quality.}}
  \label{tab:results_reconstruction_optitrack}
    \centering
    \footnotesize
    \textcolor{blue}{
    \begin{tabular}{c c c c c c c c c}
    \hline
        UAVs & Regions & Pipeline & Pose & PSNR [dB] \textcolor{red}{$\uparrow$} &  SSIM \textcolor{red}{$\uparrow$} &  LPIPS \textcolor{red}{$\downarrow$} &  HD \textcolor{red}{$\downarrow$} &  WD \textcolor{red}{$\downarrow$} \\ \hline
        1 & 2 & non-RT & SfM & 8.107 & 0.963 & 0.055 & 0.113 & 0.029 \\
        1 & 2 & non-RT & OptiTrack & 8.700 & 0.955 & 0.064 & 0.126 & 0.032 \\
        1 & 2 & near-RT & SfM & 7.924 & 0.895 & 0.197 & 0.303 & 0.036 \\
        1 & 2 & near-RT & OptiTrack & 8.189 & 0.882 & 0.191 & 0.248 & 0.034 \\
        \hline
        1 & 4 & non-RT & SfM & 7.970 & 0.956 & 0.055 & 0.112 & 0.023 \\
        1 & 4 & non-RT & OptiTrack & 7.781 & 0.950 & 0.068 & 0.136 & 0.021 \\
        1 & 4 & near-RT & SfM & 7.675 & 0.873 & 0.217 & 0.228 & 0.021 \\
        1 & 4 & near-RT & OptiTrack & 7.887 & 0.905 & 0.195 & 0.272 & 0.020 \\
        \hline
        1 & 6 & non-RT & SfM & 7.871 & 0.955 & 0.059 & 0.107 & 0.023 \\
        1 & 6 & non-RT & OptiTrack & 8.079 & 0.958 & 0.059 & 0.139 & 0.024 \\
        1 & 6 & near-RT & SfM & 7.561 & 0.864 & 0.212 & 0.272 & 0.032 \\
        1 & 6 & near-RT & OptiTrack & 7.953 & 0.890 & 0.187 & 0.277 & 0.035 \\
        \hline
        1 & 8 & non-RT & SfM & 8.066 & 0.961 & 0.063 & 0.149 & 0.025 \\
        1 & 8 & non-RT & OptiTrack & 8.345 & 0.951 & 0.072 & 0.119 & 0.028 \\
        1 & 8 & near-RT & SfM & 7.577 & 0.877 & 0.200 & 0.281 & 0.024 \\
        1 & 8 & near-RT & OptiTrack & 7.828 & 0.894 & 0.197 & 0.278 & 0.028 \\
        \hline
        2 & 4 & non-RT & SfM & 7.247 & 0.939 & 0.084 & 0.210 & 0.022 \\
        2 & 4 & non-RT & OptiTrack & 7.673 & 0.949 & 0.070 & 0.145 & 0.026 \\
        2 & 4 & near-RT & SfM & 7.892 & 0.910 & 0.184 & 0.252 & 0.032 \\
        2 & 4 & near-RT & OptiTrack & 8.017 & 0.892 & 0.189 & 0.144 & 0.031 \\
    \hline
    \end{tabular}
    }
    \vspace{-4mm}
\end{table*}

\begin{figure*}[!t]
\centering
\subfigure[PSNR by pipeline and pose]{
\includegraphics[width=0.3\textwidth]{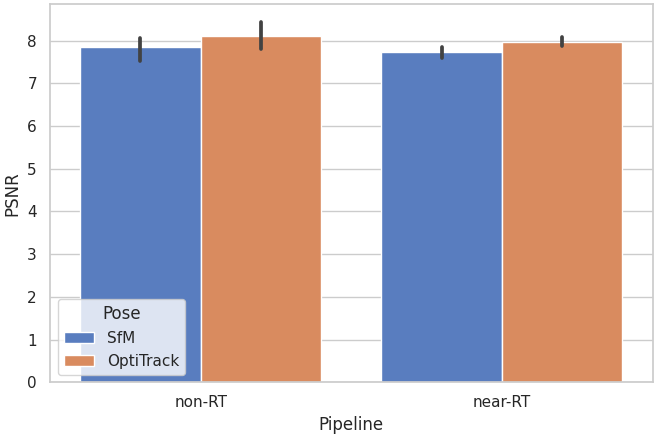}}
\subfigure[SSIM by pipeline and pose]{
\includegraphics[width=0.3\textwidth]{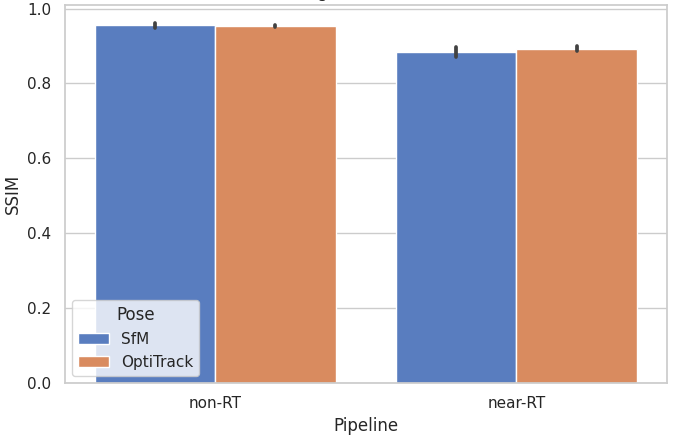}}
\subfigure[LPIPS by pipeline and pose]{
\includegraphics[width=0.3\textwidth]{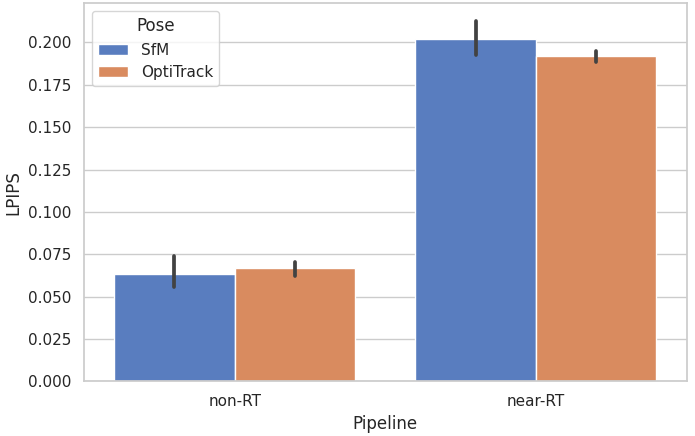}}
\vspace{-1mm}
\caption{\textcolor{blue}{Parameter sensitivity and performance ablation under OptiTrack localization, demonstrating the impact of pose fusion across non-\ac{RT} and near-\ac{RT} pipelines.}}
\vspace{-5mm}
\label{fig:param_sensitivity}
\end{figure*}

\subsection{\textcolor{blue}{Reconstruction under Constrained Image Budgets}}
\label{subsec:eval_budget}

\textcolor{blue}{To evaluate the operational efficiency of the multi-UAV setup and the dynamic trajectory adaptation independent of the total number of captured images, we conducted a sensitivity analysis based on constrained image budgets.} \textcolor{red}{This analysis was conducted offline, algorithmically subsampling from an exhaustively acquired pre-collected image pool.} \textcolor{blue}{Specifically, we evaluated the non-RT 3D reconstruction quality across fixed image counts of 50, 100, 150, 200, and 250 images. This analysis compares four configurations combining single- and dual-UAV deployments with static and dynamic adaptation, supported by Loco UWB and targeting a small object.}

\textcolor{blue}{The results in Figure~\ref{fig:plot_per_budget} demonstrate the advantages of spatial diversity and active viewpoint selection. When constrained to a low budget of 50 images, the dual-UAV dynamic configuration achieves an SSIM of 0.953 and an LPIPS of 0.141. In contrast, the single-UAV static baseline achieves an SSIM of 0.930 and an LPIPS of 0.219 under the same budget. This indicates that the dual-UAV setup captures complementary spatial perspectives more efficiently, reaching a higher reconstruction quality with fewer images.}

\textcolor{blue}{As the image budget increases, the dynamic trajectory approaches generally yield better reconstruction metrics than their static counterparts. At the maximum evaluated budget of 250 images, the dual-UAV dynamic approach achieves the highest overall performance, with a PSNR of 9.177 dB, an SSIM of 0.964, an LPIPS of 0.088, and an HD of 0.033. 
The sudden drop in PSNR, observed for the single UAV with dynamic trajectory adaptation on the~200-image dataset, can be attributed to the metric's instability. Specifically, as noted in~\cite{castillo2025neural3d}, some 2D imagery metrics, including PSNR, lack the sensitivity to evaluate small-scale 3D models, as they can experience abrupt penalties from minor rendering artifacts.}

\textcolor{blue}{Regardless, these findings indicate that the dual-UAV system does not merely rely on capturing more images to improve quality, but structurally benefits from spatial segregation and trajectory optimization to enhance sample efficiency.}

\vspace{-1mm}
\subsection{Parameter Sensitivity and Upper-Bound Benchmarking}
\label{subsec:eval_optitrack}

\textcolor{blue}{We execute a four-dimensional ablation study utilizing the high-precision OptiTrack setup, as summarized in Table~\ref{tab:results_reconstruction_optitrack} and Figure~\ref{fig:param_sensitivity}.} \textcolor{red}{This analysis was conducted offline, by replaying the exhaustively pre-collected image pool.}
\textcolor{blue}{The OptiTrack benchmark} \textcolor{red}{allows us to examine the effect of pose fusion under high-precision localization. \textcolor{red}{As shown in Table~\ref{tab:results_reconstruction_optitrack} and Figure~\ref{fig:param_sensitivity}, incorporating OptiTrack poses raises the PSNR in nine of ten configurations, whereas HD, SSIM, and LPIPS improve in only about half of them.} For example, in the single-UAV 2-Region non-RT configuration, PSNR increases from 8.107~dB to 8.700~dB. More notably, in the dual-UAV 4-Region configuration, OptiTrack fusion reduces the HD from 0.210 to 0.145, showing a substantial improvement in structural fidelity over SfM.} \textcolor{red}{However, this PSNR benefit is not absolute across all configurations. For example, in the single-UAV 4-Region non-RT setup, OptiTrack PSNR falls slightly from 7.970~dB to 7.781~dB, and HD rises from 0.112 to 0.136, indicating that pose fusion remains sensitive to spatial partitioning geometries.}

\begin{figure}
\centering
\includegraphics[width=0.37\textwidth]{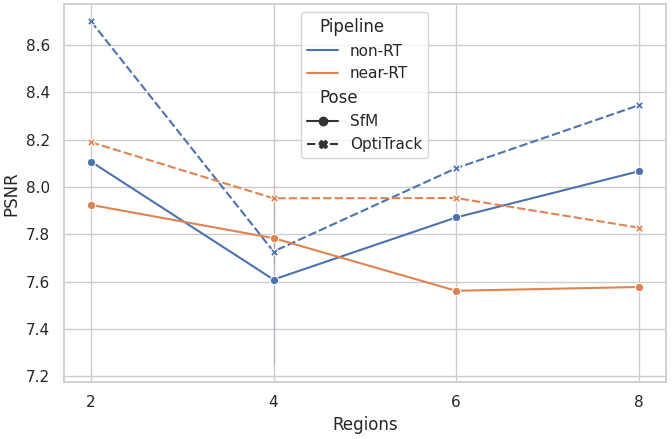}
\vspace{-1mm}
\caption{The impact of spatial partitioning (number of regions) on overall reconstruction PSNR.}
\label{fig:param_sensitivity_nr_regions}
\vspace{-4mm}
\end{figure}

\textcolor{blue}{The region sensitivity analysis (cf., Figure~\ref{fig:param_sensitivity_nr_regions}) highlights the tradeoffs inherent to spatial partitioning. While configurations such as $R=2$ or $R=8$ achieve high non-RT PSNR in single-UAV modes, $R=4$ offers a balanced tradeoff between spatial granularity, geometric accuracy (lower HD), and near-RT consistency. These experiments \textcolor{red}{indicate that high-precision location-awareness can be valuable for maximizing 3D reconstruction fidelity, depending on the spatial partitioning.}}

\vspace{-1mm}
\subsection{Reconstruction Scalability}
\label{subsec:eval_scalability}

To evaluate the scalability and robustness of the pipeline, we extend our experiments beyond the original small object by introducing a significantly more complex and larger anatomical test case. We conduct reconstructions of both objects using \ac{BW} and RGB image inputs captured with a single-\ac{UAV} configuration supported by Loco UWB localization. \textcolor{red}{These results were obtained during fully autonomous, online closed-loop flights.}
The results are derived for the integrated approach, \textcolor{red}{as it combines location-awareness and dynamic path adaptation.}

\begin{figure}[!t]
\centering
\vspace{-3mm}
\subfigure[Point-cloud]{
\includegraphics[width=0.145\textwidth]{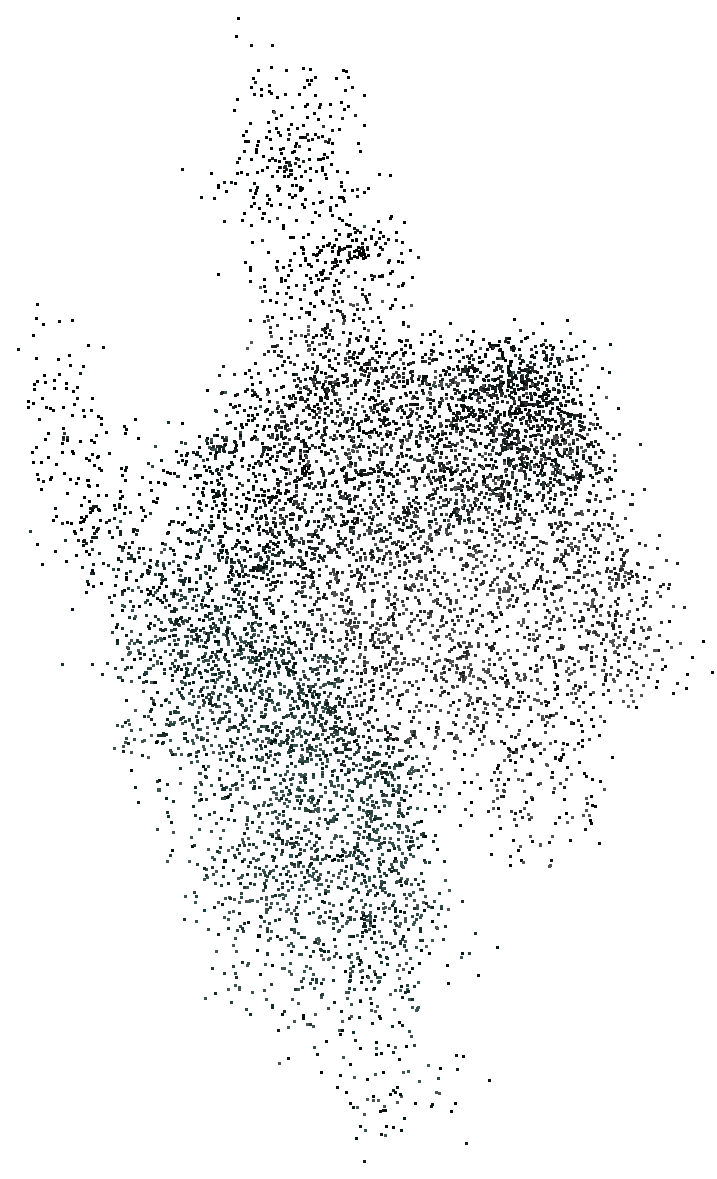}}
\subfigure[Nerfacto render, BW]{
\includegraphics[width=0.155\textwidth]{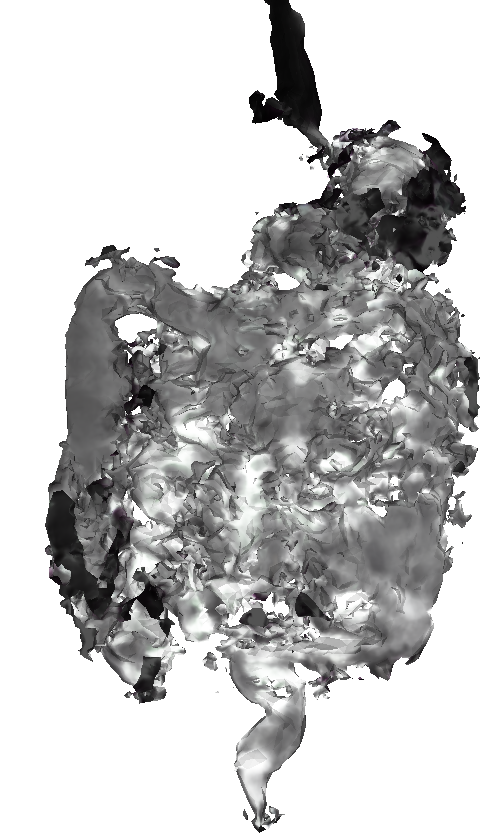}}
\subfigure[Nerfacto render, RGB]{
\includegraphics[width=0.155\textwidth]{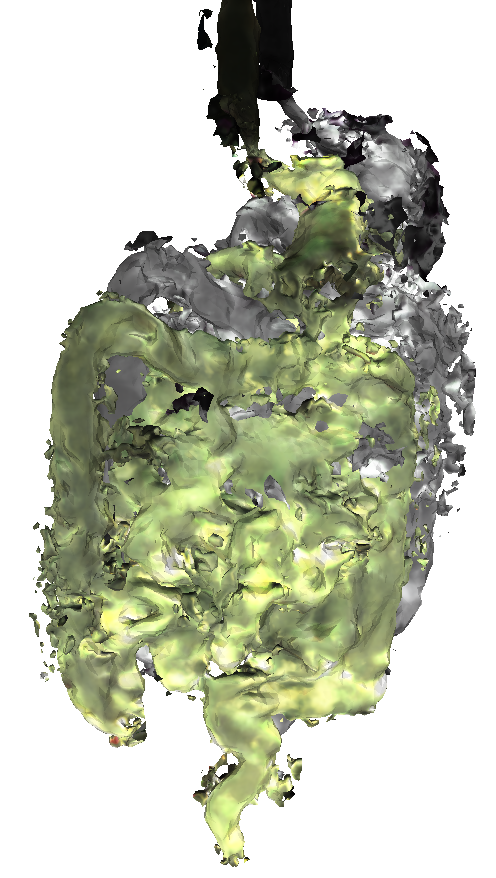}}
\vspace{-1mm}
\caption{Point-cloud and Nerfacto renders of the large (gastrointestinal) object and different camera types.}
\vspace{-3mm}
\label{fig:large_object}
\end{figure}

The results highlight that the pipeline scales effectively with both visually and structurally complex targets. For the small object, \ac{BW} reconstructions perform slightly better than RGB across both near- and non-RT pipelines. 
The corresponding SfM pointclouds and Nerfacto renders are presented in Figure~\ref{fig:large_object}.
\ac{BW} achieves higher SSIM (0.971 vs. 0.968 in near-RT, 0.984 vs. 0.978 in non-RT) and nearly identical HD values (0.126-0.127 near-RT, 0.110-0.112 non-RT), indicating marginally higher structural similarity and geometric accuracy.

For the large object, the trend reverses, with RGB \textcolor{red}{generally} outperforming \ac{BW}. In both near- and non-RT reconstructions, RGB achieves higher SSIM (0.958 vs. 0.929 in near-RT, 0.986 vs. 0.949 in non-RT) and lower HD (114.37 vs. 182.36 in near-RT, 95.23 vs. 129.04 in non-RT). This demonstrates that RGB cameras capture structural similarity more effectively while also reducing geometric deviations, particularly for complex or reflective objects where richer visual cues are necessary for stable reconstruction.

\begin{table*}[!t]
  \caption{Single-UAV system-based 3D reconstruction results for different camera types and objects.}
  \label{tab:results_scalability}
    \centering
    \footnotesize
    \begin{tabular}{l c c c c c c c}
    \hline
        \textbf{Object, Camera Type} & \makecell{\textbf{PSNR} \textcolor{red}{$\uparrow$} \\\textbf{[dB]}} &  \textbf{SSIM} \textcolor{red}{$\uparrow$} &  \textbf{LPIPS} \textcolor{red}{$\downarrow$} &  \textbf{HD} \textcolor{red}{$\downarrow$} &  \textbf{WD} \textcolor{red}{$\downarrow$} & \makecell{\textbf{\# images} \\\textbf{taken}} & \makecell{\textbf{\# images} \\\textbf{used}}  \\ \hline
        \multicolumn{8}{c}{\textbf{Near-\acl{RT} 3D reconstruction}} \\ 
        Small Object, BW & 7.859 & $0.971\pm0.018$ & $0.0549\pm0.0075$ & 0.126 & 0.026 & 291 & 291 \\
        Small Object, RGB & 6.851 & $0.968\pm0.014$ & $0.0644\pm0.0092$ & 0.127 & 0.031 & 302 & 302 \\
        Large Object, BW & 14.339 & $0.929\pm0.021$ & $0.0694\pm0.0369$ & 182.360 & 17.38 & 283 & 283 \\
        Large Object, RGB & 14.770 & $0.958\pm0.022$ & $0.0763\pm0.0422$ & 114.370 & 32.62 & 265 & 265 \\ \hline
        \multicolumn{8}{c}{\textbf{Non-\acl{RT} 3D reconstruction}} \\ 
        Small Object, BW & 7.566 & $0.984\pm0.0056$ & $0.0483\pm0.0078$ & 0.110 & 0.025 & 291 & 291 \\
        Small Object, RGB & 7.541 & $0.978\pm0.0066$ & $0.0523\pm0.0087$ & 0.112 & 0.028 & 302 & 302 \\
        Large Object, BW & 15.668 & $0.949\pm0.0226$ & $0.0667\pm0.0357$ & 129.036 & 26.88 & 283 & 283 \\
        Large Object, RGB & 16.133 & $0.986\pm0.0192$ & $0.0499\pm0.0236$ & 95.230 & 30.43 & 265 & 265 \\
        \hline
    \end{tabular}
    \vspace{-3mm}
\end{table*}

This tradeoff illustrates that image modality selection should consider both the object's material and structural complexity. While \ac{BW} images perform slightly better for compact, opaque objects, RGB images provide superior results for larger or reflective objects. These findings further validate the adaptability of our reconstruction framework across varied scenarios and underline the importance of location-aware modality selection in autonomous UAV-based scanning systems.
Table~\ref{tab:results_scalability} summarizes the quantitative results for all configurations.

\vspace{-1mm}
\subsection{Reconstruction Latency}
\label{subsec:eval_latency}

\textcolor{red}{The latency measurements in this section were obtained during fully autonomous, online closed-loop flights.} A reconstruction iteration refers to the process of updating the 3D point-cloud based on the newly acquired images from the \ac{UAV}. In our approach, new reconstruction iterations are triggered in two cases. First, if the \ac{UAV} exits a region, all images captured in that region are added to the reconstructed point-cloud (typically 45 images). These images are processed in a new iteration of the 3D reconstruction pipeline to update the point-cloud. Second, a new reconstruction iteration is triggered if the elapsed time between capturing two images exceeds a predefined time threshold \textcolor{red}{(5~s)}. \textcolor{blue}{Importantly, during these near-RT processing delays of around 25-30~sec, the \ac{UAV} does not land or halt, but maintains altitude and slow-orbits its current region until the next dynamically calculated waypoint is dispatched, ensuring continuous operation.}

\begin{figure}[!t]
\centering
\includegraphics[width=0.42\textwidth]{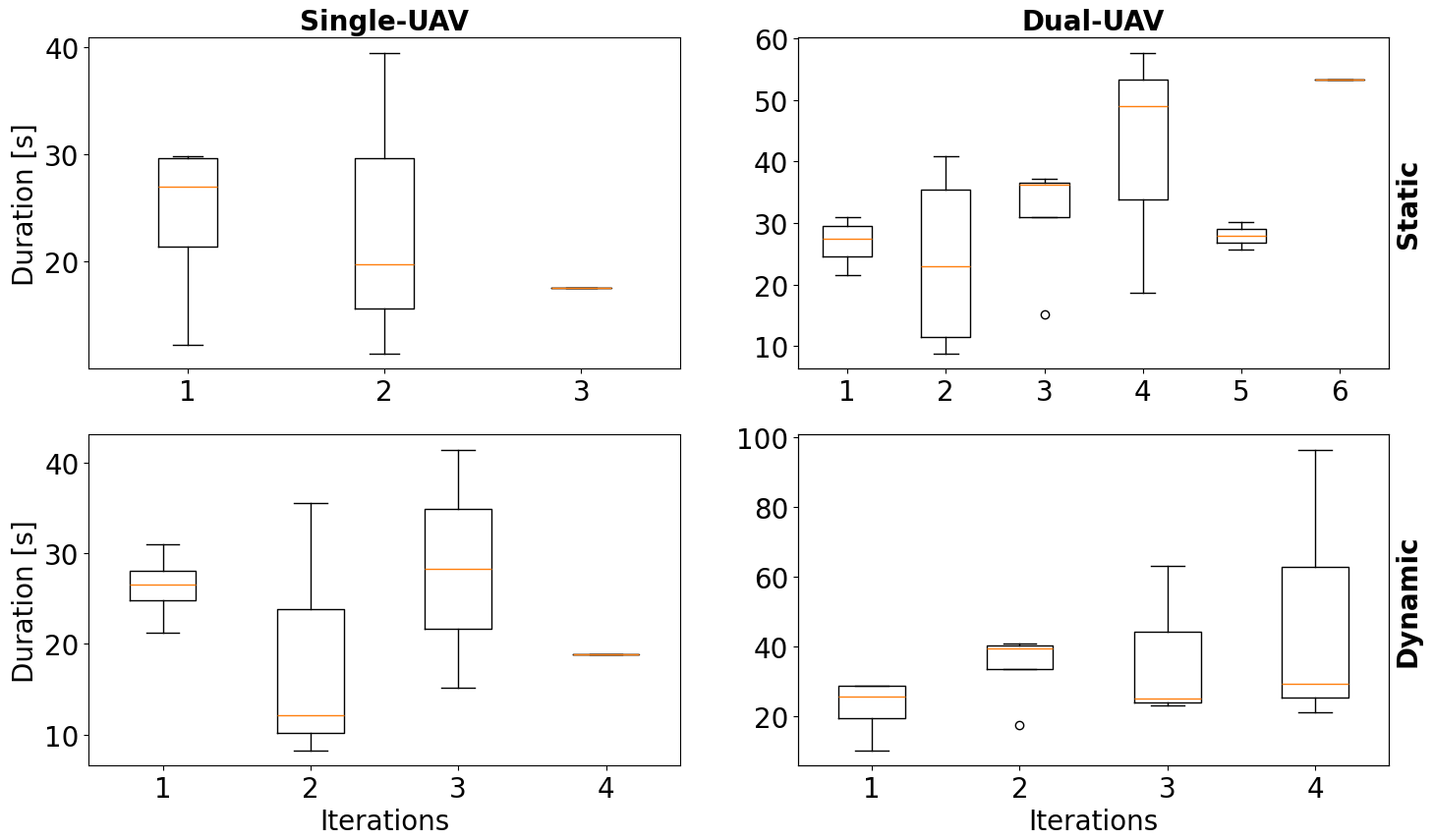}
\caption{Near-RT reconstruction latency.}
\vspace{-3mm}
\label{fig:latency}
\end{figure}

The second trigger condition improves efficiency by initiating a reconstruction process while the \ac{UAV} is moving to a new position. This was particularly necessary in cases where the movement takes longer due to a farther waypoint (in the case of dynamic trajectory) or due to imprecise position estimation with the \ac{LPS}. A reconstruction triggered by the second condition processes fewer images (as few as 2), which reduces processing time but leads to inconsistencies in near-\ac{RT} latency measurements (as seen in the high standard deviations in Tables~\ref{tab:results_reconstruction_single_uwb} and~\ref{tab:results_reconstruction_dual_uwb}, and depicted in Figure~\ref{fig:latency}). From these measurements, we conclude that while the baseline latency was lower for single-\ac{UAV} experiments (due to the smaller number of images used), in the dual-\ac{UAV} setup, the dynamic trajectory and the need for both \acp{UAV} to arrive at the waypoint resulted in more frequent triggering of the reconstruction process, leading to lower latency despite more images used. \textcolor{red}{Importantly, because the number of images processed per iteration varies significantly across approaches, directly comparing absolute iteration latency does not accurately reflect true computational efficiency. Therefore, these latency values strictly denote operational system delays. Explicit conclusions regarding comparative computational efficiency should not be drawn from these unnormalized latencies without considering metrics such as processing time per image.}

Investigating the non-\ac{RT} reconstruction latencies shows that the reconstruction with the dual-\ac{UAV} setup is generally faster than the single-\ac{UAV} counterpart. This might be considered counterintuitive, given that the latency is expected to increase with the number of images. The shorter reconstruction latency can be explained because more images provided to Nerfacto \textcolor{red}{in the dual-\ac{UAV} setup} result in fewer iterations needed for the reconstruction. \textcolor{red}{Within each \ac{UAV} setup, however, the number of iterations was kept constant to guarantee an objective comparison of the different approaches. Consequently,} the latency increases with the number of captured images in both single and dual setups, causing the dynamic path and integrated approaches to be slightly slower.

\section{Discussion and Future Efforts}
\label{sec:discussion}

Several areas for improvement emerge from our findings. 
First, the dimensioning of the number of \acp{UAV} in a particular deployment setup should be based on the target object's size for optimizing coverage and resolution. 
\textcolor{blue}{Furthermore, our systematic region ablation underscores the framework's adaptability, demonstrating that the optimal spatial partitioning inherently depends on the deployment priority, i.e., whether the focus is rapid near-\ac{RT} trajectory guidance or maximizing offline high-fidelity volumetric rendering.}
Future efforts should focus on refining these dimensions to accommodate larger objects and improving the latency of near-\ac{RT} reconstruction. 
Evaluations should be expanded to assess the feasibility and performance of our system on a larger scale.

Second, standardizing an evaluation framework as the target reconstruction objects become smaller and more detailed will be essential for objectively comparing and benchmarking different systems and approaches in \ac{UAV}-based \ac{3D} reconstruction. 
This framework should encompass metrics for accuracy, efficiency, scalability, and usability, ensuring comprehensive assessments across diverse applications and environments.
The capability of the framework in accurately capturing the performance trends and intricate differences between reconstructions should be assessed to showcase the ability of the selected reconstruction accuracy metrics to capture the inherent limits of different reconstruction approaches. 

Third, the experiments reveal the sensitivity of reconstruction performance to camera modality and object properties. While RGB images generally yield denser point-clouds, they may introduce more artifacts in the absence of surface reflections, as observed with the smaller object. In contrast, BW images, although producing sparser point-clouds, achieved more accurate reconstructions in such scenarios. 
Future designs could consider adaptive camera modality based on real-time scene analysis, e.g., preferring RGB sensors for reflective or complex geometries and BW for high-contrast objects.

Fourth, from the system's perspective, integrating features like obstacle avoidance based on near-\ac{RT} reconstruction capabilities could enhance the capabilities of our prototype. 
These advancements would improve sensing capabilities and expand the range of applications, particularly in complex and dynamic environments where real-time decision-making and obstacle detection are crucial.
\textcolor{red}{Moreover, the viewpoint selection always targets the lowest-scoring region and does not distinguish regions that remain poorly reconstructed due to missing texture from regions that lack viewpoints; revisit penalties or a score-gain-based termination criterion are directions for future work. The score also uses point counts that are normalized but not capped, and a clustering radius expressed in the \ac{SfM} units of each region, which should be normalized across regions.}

Fifth, enhancing the reliability of \ac{UAV} operations is paramount. 
\textcolor{blue}{Deploying micro-\acp{UAV} in practical scenarios exposes them to significant environmental vulnerabilities. Future system designs must explicitly account for phenomena often encountered in real deployments, such as the limited wind resistance of small \acp{UAV}, potential actuator failure, and strict limits on battery life and flight time.} This also includes an often overlooked issue of degradation in the performance of radio-based localization, such as the \ac{UWB} system utilized in this work, due to radio interference~\cite{behboodi2015interference,lemic2015experimental}, which caused hurdles in \ac{UAV} control. 
Moreover, enhancing electronic components' reliability and optimizing system-level performance will be essential for ensuring robust operation across scenarios.

Sixth, adaptations of our \ac{UAV} setup to use high-frequency \ac{JCAS}~\cite{wu2022sensing,lemic2021survey,han2024thz} for communication, 3D imaging, and \ac{UAV} localization and control. This would open new avenues for \ac{3D} reconstruction in dynamic and uncontrolled environments, as well as prolonged flight duration due to lower energy consumption of the system. 

Finally, exploring advanced volumetric rendering techniques such as \ac{GS} and instant-ngp, as well as their optimal hyperparameterizations could further improve the performance.
Our initial attempt at the utilization of Splatfacto implementation of GS is promising~\cite{castillo2025neural3d}, yet a reduction in its implementation's complexity is needed for wider-scale deployments. These approaches offer enhancements in accuracy, detail preservation, and computational efficiency, thus advancing small \ac{UAV}-based \ac{3D} reconstruction.
They will serve as an enabler of additional applications such as \ac{3D} anomaly detection.

\section{Conclusion}
\label{sec:conclusion}

We have presented a novel system architecture for utilizing small \acp{UAV} to autonomously generate high-resolution 3D digital representations of static objects. Our implemented system harnesses the capabilities of small Crazyflie \acp{UAV} as carriers and power sources for camera sensors, enabling them to capture images from multiple vantage points around the object of interest. These images are transmitted to a base station executing an open-source pipeline for near-\ac{RT} and non-\ac{RT} 3D object reconstruction.

We have experimentally demonstrated encouraging reconstruction performance of the setup for single and \textcolor{red}{dual-\ac{UAV} deployments governed by altitude and angular separation constraints}, and for small and large objects. \textcolor{blue}{Our results demonstrate that context-awareness via accurate camera localization, evaluated across both Loco \ac{UWB} radio tracking and OptiTrack optical motion-capture setups, and algorithmic dynamic path planning,} \textcolor{red}{establish a closed-loop approach between active coverage evaluation and trajectory adjustment. This approach achieves improvements in near-RT geometric error and peak signal-to-noise ratio under most tested configurations.} We further investigate the system's scalability to larger and more complex objects, showing that performance varies with camera modality and object properties, emphasizing the importance of context-aware reconstruction strategies.

While our system represents an advancement in small \ac{UAV}-based 3D reconstruction, ongoing research and development efforts will be crucial for unlocking the full potential of small \acp{UAV} in fine-grained photogrammetry, digital mapping, anomaly detection, and environmental monitoring.

\vspace{-0.9mm}
\section*{Acknowledgments}
This work was supported by the European Union’s Horizon
Europe Smart Networks and Services Joint Undertaking (grants nº 101139161 - INSTINCT and nº 101192521 - MultiX) and Departament de Recerca i Universitats under the frame of the RESILICAT project.

\renewcommand{\bibfont}{\footnotesize}
\printbibliography

@article{mozaffari2021toward,
  title={Toward 6G with connected sky: UAVs and beyond},
  author={Mozaffari, Mohammad and Lin, Xingqin and Hayes, Stephen},
  journal={IEEE Communications Magazine},
  volume={59},
  number={12},
  pages={74--80},
  year={2021},
  publisher={IEEE}
}

@article{lemic2021survey,
  title={Survey on terahertz nanocommunication and networking: A top-down perspective},
  author={Lemic, Filip and Abadal, Sergi and Tavernier, Wouter and Stroobant, Pieter and Colle, Didier and Alarc{\'o}n, Eduard and Marquez, Johann and Famaey, Jeroen},
  journal={IEEE Journal on Selected Areas in Communications},
  volume={39},
  number={6},
  pages={1506--1543},
  year={2021},
  publisher={IEEE}
}

@article{stathopoulou2019open,
  title={Open-source image-based 3D reconstruction pipelines: Review, comparison and evaluation},
  author={Stathopoulou, Elisavet Konstantina and Welponer, M and Remondino, Fabio},
  journal={Photogrammetry, Remote Sensing and Spatial Information Sciences, Volume XLII-2/W17},
  pages={331--338},
  year={2019}
}

@techreport{schenk2005introduction,
  title={Introduction to photogrammetry},
  author={Schenk, Toni},
  institution={The Ohio State University, Columbus},
  volume={106},
  year={2005}
}

@article{mohsan2023unmanned,
  title={Unmanned aerial vehicles (UAVs): Practical aspects, applications, open challenges, security issues, and future trends},
  author={Mohsan, Syed Agha Hassnain and Othman, Nawaf Qasem Hamood and Li, Yanlong and Alsharif, Mohammed H and Khan, Muhammad Asghar},
  journal={Intelligent Service Robotics},
  volume={16},
  number={1},
  pages={109--137},
  year={2023},
  publisher={Springer}
}

@article{rovira2022review,
  title={A review of AI-enabled routing protocols for UAV networks: Trends, challenges, and future outlook},
  author={Rovira-Sugranes, Arnau and Razi, Abolfazl and Afghah, Fatemeh and Chakareski, Jacob},
  journal={Ad Hoc Networks},
  volume={130},
  pages={102790},
  year={2022},
  publisher={Elsevier}
}

@article{mourtzis2021uavs,
  title={Uavs for industrial applications: Identifying challenges and opportunities from the implementation point of view},
  author={Mourtzis, Dimitris and Angelopoulos, John and Panopoulos, Nikos},
  journal={Procedia Manufacturing},
  volume={55},
  pages={183--190},
  year={2021},
  publisher={Elsevier}
}

@article{yan2021sampling,
  title={Sampling-based path planning for high-quality aerial 3D reconstruction of urban scenes},
  author={Yan, Feihu and Xia, Enyong and Li, Zhaoxin and Zhou, Zhong},
  journal={Remote Sensing},
  volume={13},
  number={5},
  pages={989},
  year={2021},
  publisher={MDPI}
}

@article{karam2022micro,
  title={Micro and macro quadcopter drones for indoor mapping to support disaster management},
  author={Karam, S. and Nex, F. and Karlsson, O. and Rydell, J. and Bilock, E. and M. Tulldahl and M. Holmberg and N. Kerle },
  journal={ISPRS Annals of the Photogrammetry, Remote Sensing and Spatial Information Sciences},
  volume={1},
  pages={203--210},
  year={2022},
  publisher={Copernicus Publications G{\"o}ttingen, Germany}
}

@article{kamencay2012improved,
  title={Improved Depth Map Estimation from Stereo Images Based on Hybrid Method},
  author={Kamencay, Patrik and Breznan, Martin and Jarina, Roman and Lukac, Peter and Zachariasova, Martina},
  journal={Radioengineering},
  volume={21},
  number={1},
  year={2012}
}

@techreport{wawrla2019applications,
  title={Applications of drones in warehouse operations},
  author={Wawrla, Lukas and Omid Maghazei and Torborn Netland},
  institution={Whitepaper. ETH Zurich, D-MTEC},
  volume={212},
  pages={1--12},
  year={2019}
}

@inproceedings{de2022rmf,
  title={Rmf-owl: A collision-tolerant flying robot for autonomous subterranean exploration},
  author={De Petris, Paolo and Nguyen, Huan and Dharmadhikari, Mihir and others},
  booktitle={Proc.~International Conference on Unmanned Aircraft Systems (ICUAS)},
  pages={536--543},
  year={2022},
  organization={IEEE}
}

@article{xu2016skeletal,
  title={Skeletal camera network embedded structure-from-motion for 3D scene reconstruction from UAV images},
  author={Xu, Zhihua and Wu, Lixin and Gerke, Markus and Wang, Ran and Yang, Huachao},
  journal={ISPRS Photogrammetry and Remote Sensing},
  volume={121},
  pages={113--127},
  year={2016},
  publisher={Elsevier}
}

@inproceedings{behboodi2015interference,
  title={Interference effect on localization solutions: signal feature perspective},
  author={Behboodi, Arash and Wirstrom, Niklas and others},
  booktitle={Proc.~IEEE Vehicular Technology Conference (VTC Spring)},
  pages={1--7},
  year={2015},
  organization={IEEE}
}

@Article{kerbl3Dgaussians,
      author       = {Kerbl, Bernhard and Kopanas, Georgios and Leimk{\"u}hler, Thomas and Drettakis, George},
      title        = {3D Gaussian Splatting for Real-Time Radiance Field Rendering},
      journal      = {ACM Transactions on Graphics},
      number       = {4},
      volume       = {42},
      pages        = {1--14},
      year         = {2023}
}

@article{muller2022instant,
  title={Instant neural graphics primitives with a multiresolution hash encoding},
  author={M{\"u}ller, Thomas and Evans, Alex and Schied, Christoph and Keller, Alexander},
  journal={ACM Transactions on Graphics},
  volume={41},
  number={4},
  pages={1--15},
  year={2022},
  publisher={ACM New York, NY, USA}
}

@article{gao2022nerf,
  title={Neural radiance fields in 3D vision: A comprehensive review},
  author={Gao, Kyle and Gao, Yina and He, Hongjie and Lu, Dening and Xu, Linlin and Li, Jonathan},
  journal={Computational Visual Media},
  year={2026},
  publisher={TUP}
}

@inproceedings{matsuki2024gaussian,
  title={Gaussian splatting slam},
  author={Matsuki, Hidenobu and Murai, Riku and Kelly, Paul HJ and Davison, Andrew J},
  booktitle={Proc.~IEEE/CVF Conference on Computer Vision and Pattern Recognition},
  pages={18039--18048},
  year={2024}
}

@inproceedings{zhang2018unreasonable,
  title={The unreasonable effectiveness of deep features as a perceptual metric},
  author={Zhang, Richard and Isola, Phillip and Efros, Alexei and others},
  booktitle={Proc.~IEEE/CVF Conference on Computer Vision and Pattern Recognition},
  pages={586--595},
  year={2018}
}

@article{han2024thz,
  title={{THz ISAC}: A physical-layer perspective of terahertz integrated sensing and communication},
  author={Han, Chong and Wu, Yongzhi and Chen, Zhi and Chen, Yi and Wang, Guangjian},
  journal={IEEE Communications Magazine},
  volume={62},
  number={2},
  pages={102--108},
  year={2024},
  publisher={IEEE}
}

@article{wu2022sensing,
  title={{Sensing integrated DFT-spread OFDM waveform and deep learning-powered receiver design for terahertz integrated sensing and communication systems}},
  author={Wu, Yongzhi and Lemic, Filip and Han, Chong and Chen, Zhi},
  journal={IEEE Transactions on Communications},
  volume={71},
  number={1},
  pages={595--610},
  year={2022},
  publisher={IEEE}
}

@inproceedings{hore2010image,
  title={{Image quality metrics: PSNR vs. SSIM}},
  author={Hore, Alain and Ziou, Djemel},
  booktitle={Proc.~IEEE International Conference on Pattern Recognition},
  pages={2366--2369},
  year={2010}
}

@inproceedings{wang2003multiscale,
  title={Multiscale structural similarity for image quality assessment},
  author={Wang, Zhou and Simoncelli, Eero P and Bovik, Alan C},
  booktitle={Proc.~Asilomar Conference on Signals, Systems \& Computers},
  volume={2},
  pages={1398--1402},
  year={2003},
  organization={IEEE}
}

@article{maboudi2023review,
  title={A review on viewpoints and path planning for UAV-based 3D reconstruction},
  author={Maboudi, Mehdi and Homaei, MohammadReza and Song, Soohwan and Malihi, Shirin and Saadatseresht, Mohammad and Markus Gerke},
  journal={IEEE Journal of Selected Topics in Applied Earth Observations and Remote Sensing},
  year={2023},
  volume={16},
  pages={5026--5048}
}

@article{mildenhall2021nerf,
  title={Nerf: Representing scenes as neural radiance fields for view synthesis},
  author={Mildenhall, Ben and Srinivasan, Pratul P and Tancik, Matthew and Barron, Jonathan T and Ravi Ramaoorthi and Ren Ng},
  journal={Communications of the ACM},
  volume={65},
  number={1},
  pages={99--106},
  year={2021},
  publisher={ACM New York, NY, USA}
}

@article{lienard2016embedded,
  title={Embedded, real-time UAV control for improved, image-based 3D scene reconstruction},
  author={Li{\'e}nard, Jean and Vogs, Andre and Gatziolis, Demetrios and Strigul, Nikolay},
  journal={Measurement},
  volume={81},
  pages={264--269},
  year={2016},
  publisher={Elsevier}
}

@article{floreano2015science,
  title={Science, technology and the future of small autonomous drones},
  author={Floreano, Dario and Wood, Robert J},
  journal={Nature},
  volume={521},
  number={7553},
  pages={460--466},
  year={2015},
  publisher={Nature Publishing Group UK London}
}

@inproceedings{castillo2025neural3d,
  author    = {G. Castillo G{\'o}mez-Raya and {\'A}. Veres-Vit{\'a}lyos and F. Lemic and P. Royo and M. Montagud and S. Fern{\'a}ndez and S. Abadal and X. Costa-P{\'e}rez},
  title     = {Experimental Assessment of Neural 3D Reconstruction for Small UAV-Based Applications},
  booktitle = {Proc.~IEEE International Symposium on Personal, Indoor and Mobile Radio Communications (PIMRC)},
  year      = {2025}
}

@article{bugelnig2026adaptive,
  title={Adaptive Flight Planning for Neural 3D Reconstruction With Low-Altitude UAVs},
  author={Bugelnig, Daniel Johannes and Lemic, Filip and Veres-Vit{\'a}lyos, {\'A}lmos and G{\'o}mez-Raya, Gen{\'\i}s Castillo and Costa-P{\'e}rez, Xavier and Rinner, Bernhard},
  journal={IEEE Internet of Things Magazine},
  year={2026},
  publisher={IEEE}
}

@inproceedings{talarn2023real,
  title={Real-time Generation of 3-Dimensional Representations of Static Objects using Small Unmanned Aerial Vehicles},
  author={Talarn, Pau and Oll{\'e}, Bernat and Lemic, Filip and Abadal, Sergi and Costa-Perez, Xavier},
  booktitle={Proc.~ACM Mobile Computing and Networking (MobiCom)},
  pages={1--3},
  year={2023}
}

@article{gordan2021brief,
  title={A brief overview and future perspective of unmanned aerial systems for in-service structural health monitoring},
  author={Gordan, Meisam and Ismail, Zubaidah and Ghaedi, Khaled and Ibrahim, Zainah and Hashim, Huzaifa and Ghayeb, Haider Hamad and Talebkhah, Marieh},
  journal={Engineering Advances},
  volume={1},
  number={1},
  pages={9--15},
  year={2021}
}

@article{campana2017drones,
  title={Drones in Archaeology. SotA and Future Perspectives},
  author={Campana, Stefano},
  journal={Archaeological Prospection},
  volume={24},
  number={4},
  pages={275--296},
  year={2017},
  publisher={Wiley Online Library}
}

@article{liu2022challenges,
  title={Challenges and opportunities for autonomous micro-uavs in precision agriculture},
  author={Liu, Xu and Chen, Steven W and Nardari, Guilherme V and Qu, Chao and Cladera, Fernando and others},
  journal={IEEE Micro},
  volume={42},
  number={1},
  pages={61--68},
  year={2022},
  publisher={IEEE}
}

@inproceedings{lemic2015experimental,
  title={{Experimental evaluation of RF-based indoor localization algorithms under RF interference}},
  author={Lemic, Filip and Handziski, Vlado and Wolisz, Adam and Constambeys, Timotheos and Laoudias, Christos and Adler, Stephan and Schmitt, Simon and Yang, Yuan},
  booktitle={Proc.~International Conference on Localization and GNSS (ICL-GNSS)},
  pages={1--8},
  year={2015},
  organization={IEEE}
}

@article{giordan2020use,
  title={The use of UAVs for engineering geology applications},
  author={Giordan, Daniele and Adams, Marc S and Aicardi, Irene and Alicandro, Maria and Allasia, Paolo and Baldo, Marco and De Berardinis, Pierluigi and Dominici, Donatella and Godone, Danilo and Hobbs, Peter and Veronika Lechner and Tomasz Niedzielski and Marco Piras and Marianna Rotilio and Riccardo Salvini and Valerio Segor and Bernadette Sotier and Fabrizio Troilo },
  journal={Bulletin of Engineering Geology and the Environment},
  volume={79},
  pages={3437--3481},
  year={2020},
  publisher={Springer}
}

@article{kovanivc2023review,
  title={Review of photogrammetric and Lidar applications of UAV},
  author={Kovani{\v{c}}, Ludov{\'\i}t and Topitzer, Branislav and Petovsk{\`y}, Patrik and Bli{\v{s}}tan, Peter and Gergelov{\'a}, Marcela Bindz{\'a}rov{\'a} and Bli{\v{s}}tanov{\'a}, Monika},
  journal={Applied Sciences},
  volume={13},
  number={11},
  pages={6732},
  year={2023},
  publisher={MDPI}
}

@article{gao2023uav,
  title={A UAV-based explore-then-exploit system for autonomous indoor facility inspection and scene reconstruction},
  author={Gao, Chuanxiang and Wang, Xinyi and Wang, Ruoyu and Zhao, Zuoquan and Zhai, Yu and Chen, Xi and Chen, Ben M},
  journal={Automation in Construction},
  volume={148},
  pages={104753},
  year={2023},
  publisher={Elsevier}
}

@article{huang2020fast,
  title={Fast reconstruction of 3D point cloud model using visual SLAM on embedded UAV development platform},
  author={Huang, Fang and Yang, Hao and Tan, Xicheng and Peng, Shuying and Jian Tao and Siyuan Peng},
  journal={Remote Sensing},
  volume={12},
  number={20},
  pages={3308},
  year={2020},
  publisher={MDPI}
}

@article{koch2019automatic,
  title={Automatic and semantically-aware 3D UAV flight planning for image-based 3D reconstruction},
  author={Koch, Tobias and K{\"o}rner, Marco and Fraundorfer, Friedrich},
  journal={Remote Sensing},
  volume={11},
  number={13},
  pages={1550},
  year={2019},
  publisher={MDPI}
}

@article{krul2021visual,
  title={Visual SLAM for indoor livestock and farming using a small drone with a monocular camera: A feasibility study},
  author={Krul, Sander and Pantos, Christos and Frangulea, Mihai and Valente, Jo{\~a}o},
  journal={Drones},
  volume={5},
  number={2},
  pages={41},
  year={2021},
  publisher={MDPI}
}

@inproceedings{poissond2006Kazhdan,
  title={Poisson surface reconstruction},
  author={Kazhdan, Michael and Bolitho, Matthew and Hoppe, Hugues},
  booktitle={Proc.~Eurographics Symposium on Geometry Processing},
  volume={7},
  number={4},
  year={2006}
}

@inproceedings{nerfstudio,
  title={Nerfstudio: A modular framework for neural radiance field development},
author={Tancik, Matthew and Weber, Ethan and Ng, Evonne and Li, Ruilong and Yi, Brent and Wang, Terrance and Kristoffersen, Alexander and Austin, Jake and Salahi, Kamyar and Ahuja, Abhik and others},
  booktitle={Proc.~ACM Special Interest Group on Computer Graphics (SIGGRAPH)},
  pages={1--12},
  year={2023}
}

@inproceedings{alicevision2021,
  title={{A}liceVision {M}eshroom: An open-source {3D} reconstruction pipeline},
  author={Carsten Griwodz and Simone Gasparini and Lilian Calvet and Pierre Gurdjos and Fabien Castan and Benoit Maujean and Gregoire De Lillo and Yann Lanthony},
  booktitle={Proc.~ACM Multimedia Systems Conference (MMSys'21)},
  publisher = {ACM Press},
  year = {2021}
}

@ARTICLE{Rinner_Computer2021,
  author   = {Rinner, Bernhard and Bettstetter, Christian and Hellwagner, Hermann and Weiss, Stephan},
  journal  = {Computer}, 
  title    = {Multidrone Systems: More Than the Sum of the Parts}, 
  year     = {2021},
  volume   = {54},
  number   = {5},
  pages    = {34-43}
}

@article{pham2026colmar,
  title={COLMAR: Cooperative View Policy Learning for Multi-Agent Active 3D Reconstruction},
  author={Pham, Phu and Conover, Damon and Bera, Aniket},
  journal={arXiv preprint arXiv:2607.13524},
  year={2026}
}

@article{dursun2026object,
  title={Object-Reconstruction-Aware Whole-body Control of Mobile Manipulators},
  author={Dursun, Fatih and Adorno, Bruno Vilhena and Watson, Simon and Pan, Wei},
  journal={IEEE Transactions on Robotics},
  year={2026},
  publisher={IEEE}
}

@article{farchione2026less,
  title={Less Is More: Genetic Frame Selection for Efficient Novel View Synthesis},
  author={Farchione, Diego E and Idoughi, Ramzi and Jaspe-Villanueva, Alberto and Wonka, Peter},
  journal={arXiv preprint arXiv:2609.35573},
  year={2026}
}

@article{liu2021active,
  title={Active and interactive mapping with dynamic Gaussian process implicit surfaces for mobile manipulators},
  author={Liu, Liyang and Fryc, Simon and Wu, Lan and Vu, Thanh Long and Paul, Gavin and Vidal-Calleja, Teresa},
  journal={IEEE Robotics and Automation Letters},
  volume={6},
  number={2},
  pages={3679--3686},
  year={2021},
  publisher={IEEE}
}

\end{document}